\pdfoutput=1

\documentclass[11pt]{article}

\usepackage{coling}

\usepackage{times}
\usepackage{latexsym}

\usepackage[T1]{fontenc}

\usepackage[utf8]{inputenc}

\usepackage{microtype}

\usepackage{inconsolata}

\usepackage{graphicx}

\usepackage{todonotes}

\usepackage{booktabs}
\usepackage{graphicx}
\usepackage{subcaption}

\usepackage{multirow}
\usepackage[most]{tcolorbox}
\usepackage{wrapfig}
\usepackage[capitalize, nameinlink]{cleveref}
\usepackage{enumitem}
\usepackage{amsmath}
\usepackage{amssymb}
\usepackage{enumitem}

\crefname{figure}{fig.}{figs.}%
\crefname{equation}{eqn.}{eqns.}%
\crefname{table}{tab.}{tabs.}%
\crefname{section}{\S\!}{\S\!}
\crefname{subsection}{\S\!}{\S\!}
\crefname{subsubsection}{\S\!}{\S\!}
\crefname{appendix}{\S\!}{\S\!}

\title{Efficient Solutions For An Intriguing Failure of LLMs: Long Context Window Does Not Mean LLMs Can Analyze Long Sequences Flawlessly}

\author{
  Peyman Hosseini\textsuperscript{1} $\quad$
  Ignacio Castro\textsuperscript{1} $\quad$
  Iacopo Ghinassi\textsuperscript{1} $\quad$
  Matthew Purver\textsuperscript{1,2} \\
  \textsuperscript{1}School of EECS, Queen Mary University of London, London, UK \\
  \textsuperscript{2}Department of Knowledge Technologies, Jožef Stefan Institute, Ljubljana, Slovenia \\
  \{s.hosseini, i.castro, i.ghinassi, m.purver\}@qmul.ac.uk \\
}

\begin{document}
\maketitle
\begin{abstract}
Large Language Models (LLMs) have demonstrated remarkable capabilities in comprehending and analyzing lengthy sequential inputs, owing to their extensive context windows that allow processing millions of tokens in a single forward pass. However, this paper uncovers a surprising limitation: LLMs fall short when handling long input sequences. We investigate this issue using three datasets and two tasks (sentiment analysis and news categorization) across various LLMs, including Claude 3, Gemini Pro, GPT 3.5 Turbo, Llama 3 Instruct, and Mistral Instruct models. To address this limitation, we propose and evaluate ad-hoc solutions that substantially enhance LLMs' performance on long input sequences by up to 50\%, while reducing API cost and latency by up to 93\% and 50\%, respectively.
\end{abstract}

\section{Introduction}
\label{sec: Introduction}
LLMs have demonstrated remarkable capabilities in natural language understanding and generation tasks. Leveraging extensive pretraining on massive text corpora, the new generation of LLMs can perform a wide range of language tasks with minimal task-specific fine-tuning. Additionally, these LLMs are equipped with behemothic context windows that enable them to analyze inputs spanning up to tens or hundreds of pages in one forward pass. In this paper, we study the performance of Claude 3 Haiku \cite{claude3haiku20240307}, GPT3.5-Turbo \cite{gpt3.5turbo}, Gemini-1.0-pro \cite{Gemini2023}, Llama 3 8b Instruct \cite{llama3}, and Mistral 7b Instruct \cite{mistral}. Some of these LLMs are equipped with context windows that can support up to 200,000 tokens in one forward pass. 

\paragraph{Related Work.} Prompting strategies have emerged as a promising avenue for improving LLM performance by providing concise and informative input \citep{Liu+2023,Brown+2020, Jiang+23, ge+24}. These strategies involve extracting key information from the input text and presenting it to the LLM in a structured manner. However, despite being equipped with context windows capable in theory of supporting large amounts of text, the performance of LLMs often suffers on lengthy input sequences as the prompt length grows \cite{Li+2023,Li+2024}. 


On the other hand, many general summarization techniques are available for condensing lengthy texts into more manageable snippets. Extractive Summarization methods such as TextRank \cite{mihalceaT2004} are widely used to identify and extract the most significant sentences from a document for different purposes \cite{CacholaLCW2020, FengFQ22, BalcerzakJW2014, Wang+2020}. Although not designed for prompt compression, these techniques might therefore be useful in this context, and have relatively low computational overheads; in this paper, we therefore investigate the use of real-time summarization pipelines and text truncation techniques to boost LLM performance by optimizing the input while reducing their load.

\begin{figure*}[th]
    \centering

    \begin{subfigure}[b]{0.47\textwidth}
        \centering
        \includegraphics[width=\textwidth]{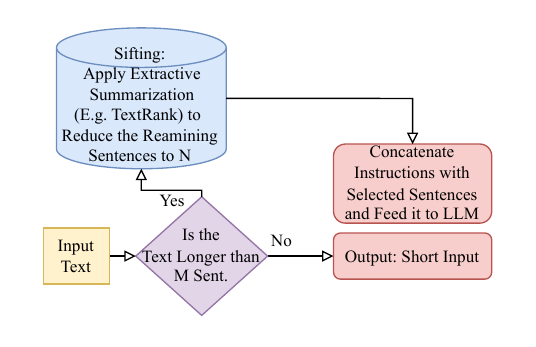}
        \caption{Pure Extractive Summarization Pipeline}
        \label{fig: Simple Summary Schemes}
    \end{subfigure}
    \hfill
    \begin{subfigure}[b]{0.5\textwidth}
        \centering
        \includegraphics[width=\textwidth]{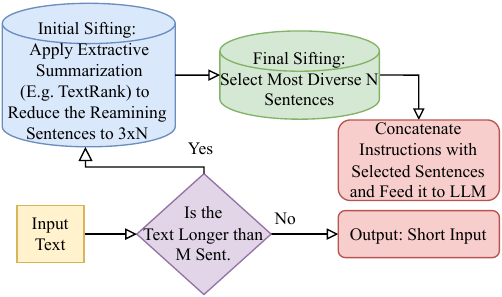}
        \caption{Diverse Summarization Pipeline}
        \label{fig: Diverse Summary Schemes}
    \end{subfigure}
    \caption{The summarization pipelines for summarising information. The diverse summarization approach builds on top of the purely extractive approach but gives higher priority to lexical diversity.}
\end{figure*}

\paragraph{Motivation.} 
There is a body of research dedicated to studying the limitations of LLMs on long sequences and proposing mitigations at both architecture-level \cite{BeltagyPC2020,BertschANG2024} as well as prompt-level \cite{Wei+2022}. These studies often involve defining and exploring overly complex problems such as those about extreme-label classification \cite{Li+2024} or ``Needle In a Haystack" \cite{MachlabB2024}. However, a systematic study of LLM capabilities and limitations on long-form analysis tasks such as news categorization or sentiment analysis
of long reviews which require a common general understanding of the input context is still lacking. Furthermore, the emphasis on approaches involving prompt-tuning has diverted attention away from optimizing and streamlining the information fed to LLMs. This study serves to fill these gaps by showcasing the failure of LLMs on canonical NLP tasks when dealing with long sequences and to ignite a spark of interest in the research community to explore the untapped potential of optimizing and condensing the information fed to LLMs.

\paragraph{Contribution.} Our main contributions are:
\begin{enumerate}[itemsep=0pt, topsep=0pt, partopsep=0pt, leftmargin=*]
\item We systematically study the performance of state-of-the-art (SotA) LLMs on sentiment analysis and news categorization, revealing their limitations in processing long-form text effectively.
\item We propose and evaluate ad-hoc solutions using extractive and diverse summarization as well as selective truncation to condense input text, which substantially improves LLM performance by up to 50\%, reduces API costs by as much as 93\% and significantly reduces latency.
\item We present comprehensive empirical and ablation studies examining the relationship between input length, summarization strategies, and model performance, providing insights into optimal summarization approaches for LLMs.
\end{enumerate}

\section{Methodology}
\label{sec: Theory}

\begin{table*}[th!]
  \smallskip
  \resizebox{0.99\textwidth}{!}{%
  \begin{tabular}{cccccc}
    \toprule
    Dataset & Domain & \#Classes & Language & Min. Thresh. (\#sent.) & Avg. Inp. Len. (\#tokens)  \\
    \midrule
    GameSpot & Game Reviews & 10\textsuperscript{*} & English & 45 & 2120 \\
    BBC News & News Documents & 5 & English & 35 & 1150 \\
    20 NewsGroup & News Documents & 20 & English & 60 & 3450 \\
    \bottomrule
  \end{tabular}
  }
  \caption{Datasets' key information. ``Min. Thresh''. indicates the minimum sentence count used to filter documents suitable for our study, ensuring only sufficiently long documents are included. ``Avg. Inp. Len'' shows the mean document length in the studied subset for each dataset. For GameSpot, review scores (originally 1-100) are rounded to the nearest multiple of 10, resulting in 10 classes.}
  \label{tbl: Datasets Summary}
\end{table*}

First, we present our summarization pipelines; then our evaluation scenarios. 

\subsection{Summarization Methodology}
We study two pipelines for extracting key information from the documents and providing input for prompting. Given our requirement for real-time operation and semantic fidelity, we investigate only extractive approaches here.

\paragraph{Pure Extractive Summarization:} As shown in \Cref{fig: Simple Summary Schemes}, we use TextRank \citep{mihalceaT2004}, a well-known unsupervised extractive summarization algorithm, to select the most important sentences. TextRank uses a graph-based ranking model to measure the similarity between sentences and their centrality within the graph. We then write the instruction to the LLM (i.e., categorize or rate) and append the extracted sentences as input.
\paragraph{Diverse Summarization:} Builds on top of the previous approach. We discard the least relevant sentences using TextRank ranking. Then we use TF-IDF to represent the sentences as vectors and calculate the diversity scores based on the dissimilarity between sentences using cosine similarity. The top N sentences with the highest diversity scores are chosen as the input used in prompting (see \Cref{app: diverse summarization}). This extension aims to maximise the diversity of the information for the LLMs.

\subsection{Prompting Scenarios}
\label{sec: Methodology sub prompting}
To investigate the performance of LLMs on sentiment analysis and news categorization tasks involving 
long 
input sequences, we employ 7 prompting strategies and evaluate their effectiveness on three datasets, which we introduce in the next section. These prompting strategies include:

\begin{enumerate}[itemsep=0pt, topsep=0pt, partopsep=0pt, leftmargin=*]
\item \textbf{Full Context:} The entire lengthy review is provided as input for analysis (Motivation: the baseline approach for comparison with other methods).
\item \textbf{Full Context + Summary:} The $N$-sentence summary extracted using \Cref{fig: Simple Summary Schemes} pipeline is appended to the lengthy review. (Motivation: how does emphasizing a selected summary with repetition affect the performance?)
\item \textbf{First Sentences:} We crop the initial $N$ sentences from the text and provide it as input. (Motivation: how does choosing the `opening' section of a lengthy review affect the performance?)
\item \textbf{Last Sentences:} We crop the ending $N$ sentences from the text and provide it as input. (Motivation: how does choosing the `ending' section of a lengthy review affect the performance?)
\item \textbf{Summary:} We provide the extracted $N$ sentence summary (\Cref{fig: Simple Summary Schemes}) as input. (Motivation:  how does choosing a summary affect performance?)
\item \textbf{Diverse Summary:} We provide the extracted $N$-sentence summary (\Cref{fig: Diverse Summary Schemes}) as input. (Motivation: how does giving more priority to lexical diversity in the summary affect performance?)
\item \textbf{Random Sampling:} We randomly select $N$ sentences from the document. (Motivation: how does randomly choosing a short snippet perform in comparison to providing the full context?) 
\end{enumerate}





%
\section{Evaluation}
\label{sec: Evaluation}
\subsection{Datasets}
\label{sec: Datasets}

We now introduce the datasets we study (see \Cref{sec:appendix} for more details). For each dataset, as our interest is in LLM performance with long inputs, we 
use only
the subsets of the data that exceed a minimum length. 
We report the average length of the studied subset (in number of tokens) in \Cref{tbl: GameSpot,tbl: 20NewsGroup,tbl: BBC News}. Additionally, \Cref{tbl: Datasets Summary} aggregates and reports key information about each dataset in a nutshell.
\paragraph{GameSpot Reviews~\cite{GameSpotReviews}:} 
more than 12,000 long game reviews with a sentiment score assigned by the author ranging from 1 to 100. 

\paragraph{20 Newsgroups~\cite{20NewsGroupDataset}:} 
nearly 20,000 news documents belonging to 20 different 
topic categories.
High-level topics include \textit{politics}, \textit{religion}, \textit{sports}, and \textit{computers}. 

\paragraph{BBC News Archive~\cite{BBCNewsDataset}:} 
2,225 BBC articles covering  \textit{business}, \textit{entertainment}, \textit{politics}, \textit{sport}, and \textit{tech}.


\subsection{Experiments}
\label{subsec: Experiments}
We evaluated the performance of Claude 3 Haiku, Gemini-1.0-Pro, GPT-3.5 Turbo, Llama 3 8b Instruct, and Mistral 7b Instruct on the datasets and tasks 
detailed in \Cref{sec: Datasets}, 
using
the prompting scenarios discussed in \Cref{sec: Methodology sub prompting}. The summarized results 
are available in \Cref{tbl: GameSpot Summary,tbl: 20NewsGroup Summary,tbl: BBC News Summary}. More detailed analysis for each LLM is available in \Cref{tbl: GameSpot,tbl: 20NewsGroup,tbl: BBC News} in 
\Cref{sec:appendix}. Aside from major performance metrics (i.e., loss, accuracy, and F1), we also report the average latency for each scenario under `Avg. Lat.' column and the average input length (in terms of tokens) under `Inp. Len.' column in all tables.

Please also note that the parameter N (discussed in items 2-7 of \Cref{sec: Methodology sub prompting}) in \Cref{tbl: GameSpot Summary,tbl: 20NewsGroup Summary,tbl: BBC News Summary} for all summarization and truncation scenarios is set to 7. We chose 7 sentences for the summary/truncation length using insights driven from our ablation study presented in \Cref{fig: Avg performance gamespot}. That is 7 sentences provide a sweet spot between short length and performance.

\subsubsection{Sentiment Analysis}

\begin{table}[th]
  \smallskip
  \resizebox{0.49\textwidth}{!}{%
  \begin{tabular}{lrrrrr}
    \toprule
    Scenario & MSE & MAE & Accuracy & Avg. Lat. & Inp. Len. \\
    \midrule

    Full & 272.8 (6) & 11.7 (6) & 36.0 (7) & 1.27 (6) & 2120 (6) \\

    Full+Sum. & 403.1 (7) & 13.5 (7) & 38.2 (6) & 1.33 (7) & 2450 (7) \\

    First Sent. & 169.1 (5) & 9.9 (5) & 41.2 (5) & 0.82 (1) & 320 (1) \\

    Last Sent. & \textbf{99.6 (1)} & \textbf{7.9 (1)} & 50.7 (2) & 0.82 (1) & 320 (1) \\

    Sum. & 124.2 (2) & 8.8 (4) & 48.2 (4) & 0.82 (1) & 320 (1) \\

    Div. Sum. & 133.7 (4) & 8.6 (2) & \textbf{54.0 (1)} & 0.82 (1) & 320 (1) \\

    Rand. Samp. & 129.6 (3) & 8.7 (3) & 50.0 (3) & 0.82 (1) & 320 (1) \\
    \bottomrule
  \end{tabular}%
  }
  \caption{Average performance of different LLMs for Sentiment Analysis on GameSpot over 5 runs.}
  \label{tbl: GameSpot Summary}
\end{table}

As shown in \Cref{tbl: GameSpot Summary}, both Full and Full+Sum approaches failed to perform favourably in predicting the article scores. However, extracting a subset of the input text, and providing it in the prompt, even through randomly sampling sentences, yielded superior performance.  The `Last Sent.' scenario performs the best in both loss metrics while the `Div Sum.' achieves the highest accuracy by a substantial margin, performing 50\% better than when providing the LLM with Full Context. Detailed analysis of the performance for each LLM on the GameSpot dataset is available in \Cref{tbl: GameSpot}.


\subsubsection{News Categorization}
We evaluated the performance of LLMs on two news categorization datasets. \Cref{tbl: 20NewsGroup Summary,tbl: BBC News Summary} summarize these results across different prompting scenarios. Detailed analyses for each LLM for both experiments are available in \Cref{tbl: 20NewsGroup,tbl: BBC News} in
\Cref{sec:appendix}. 
\begin{figure*}[ht]
    \centering
    \begin{subfigure}[b]{\textwidth}
        \centering
        \includegraphics[width=\textwidth]{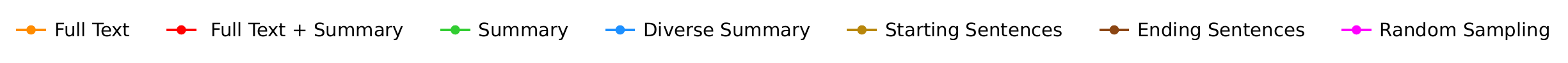}
    \end{subfigure}

    \vspace{-0.5em}
    
    \begin{subfigure}[b]{0.328\textwidth}
        \centering
        \includegraphics[width=1.05\textwidth]{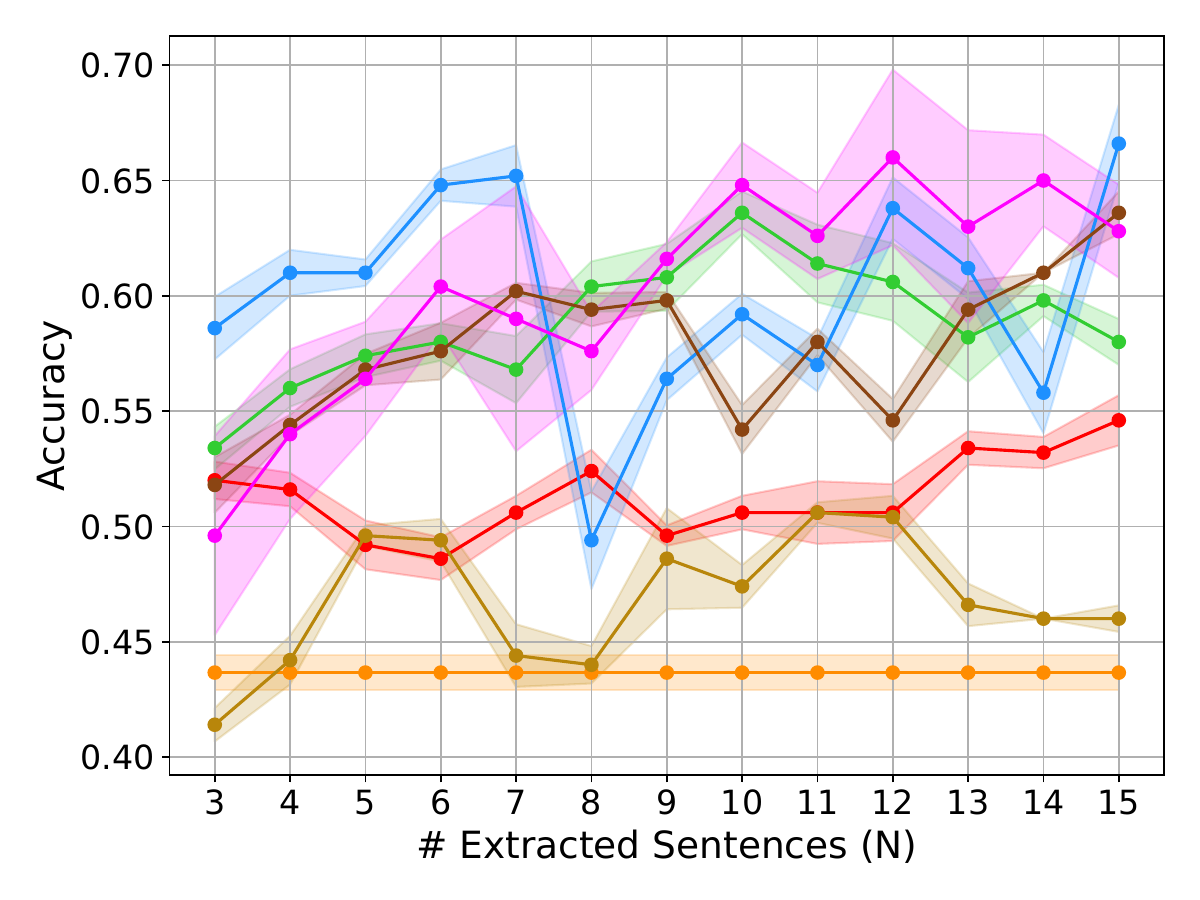}
        \label{fig: Avg Acc claude-gamespot}
    \end{subfigure}
    \hfill
    \begin{subfigure}[b]{0.328\textwidth}
        \centering
        \includegraphics[width=1.05\textwidth]{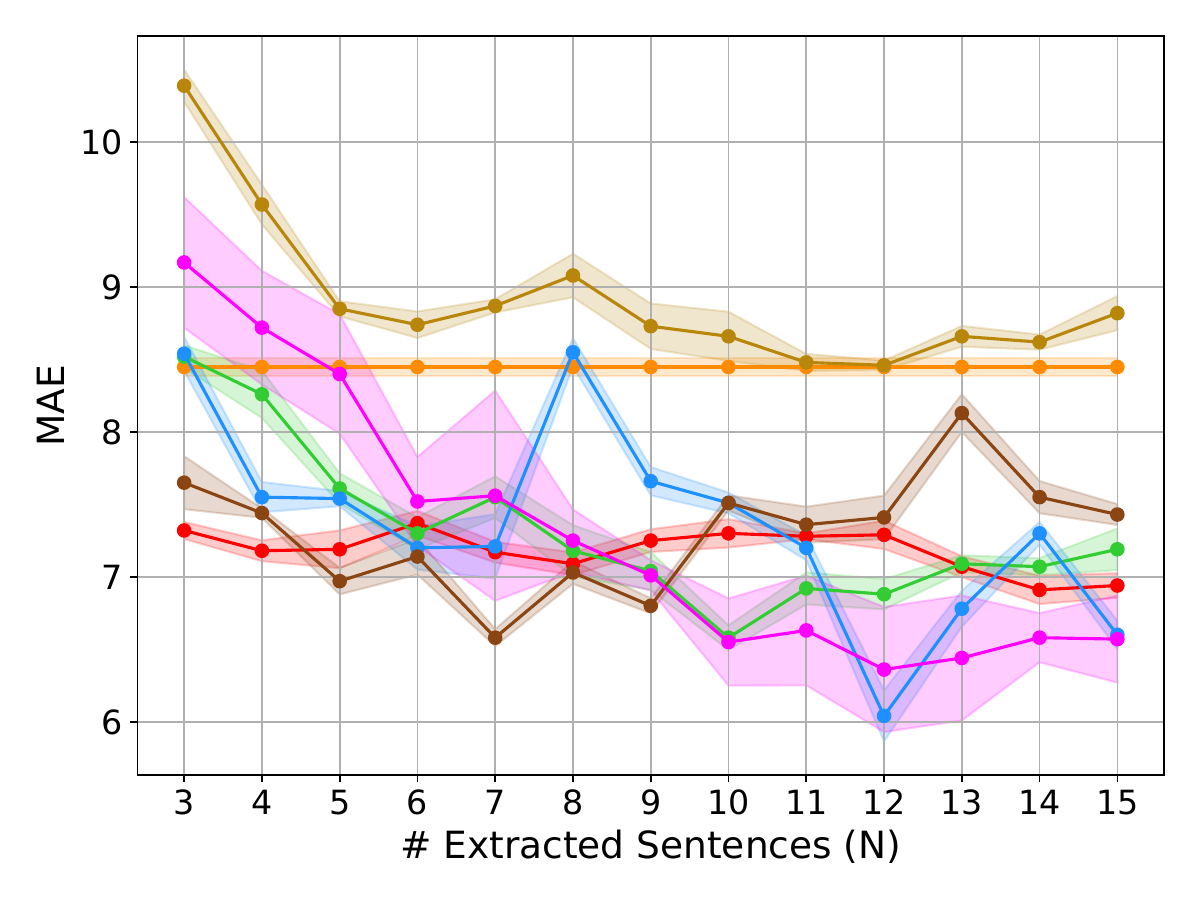}
        \label{fig: Avg MAE claude-gamespot}
    \end{subfigure}
    \hfill
    \begin{subfigure}[b]{0.328\textwidth}
        \centering
        \includegraphics[width=1.05\textwidth]{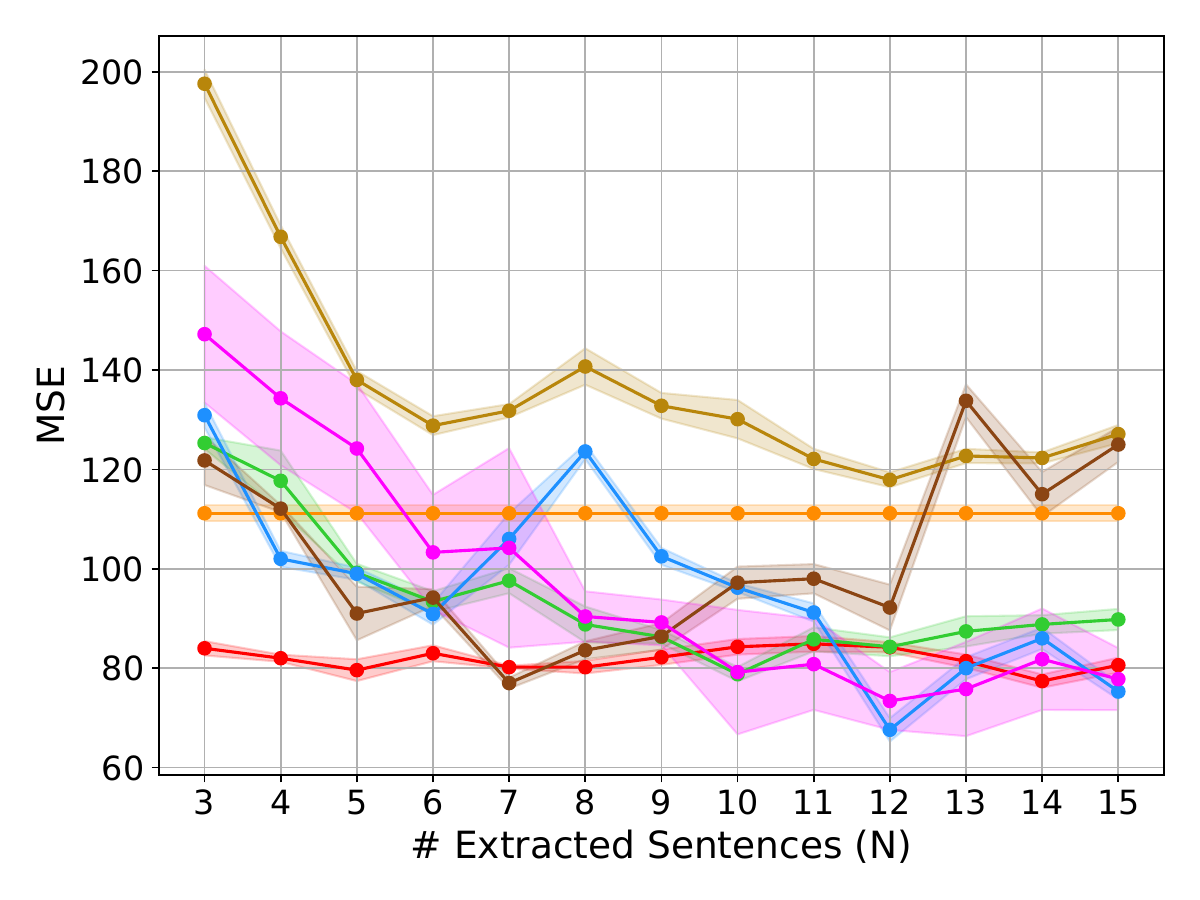}
        \label{fig: Avg MSE claude-gamespot}
    \end{subfigure}
    \vspace{-3.0em}
    \caption{Ablation study on the length of the selected truncation/summary for different scenarios using Claude 3 Haiku over 5 runs with 85\% Confidence Intervals. The results show the efficacy of approaches optimizing LLMs' input. `Full' context performs poorly on all metrics. Additionally, after the length of input exceeds 10 sentences, less meaningful improvement in the performance of all scenarios is observed. A similar trend is seen for all LLMs.}
    \label{fig: Claude GameSpot Performance}
    \vspace{-0.7em}
\end{figure*}

\begin{table}[th]
  \vspace{-0.3em}
  \smallskip
  \resizebox{0.49\textwidth}{!}{%
  \begin{tabular}{crrrr}
    \toprule
    Scenario & Mac. F1 & Acc.  & Avg. Lat. & Inp. Len.  \\
    \midrule

    Full & 0.27 (7) & 35.2 (7) & 1.58 (6) & 3450 (6) \\

    Full+Sum. & 0.30 (3) & 38.2 (5) & 1.94 (7) & 3700 (7) \\

    First Sent. & 0.30 (3) & 39.1 (3) & 0.79 (1) & 240 (1) \\

    Last Sent. & 0.29 (5) & 39.1 (3) & 0.79 (1) & 240 (1) \\

    Sum. & \textbf{0.31 (1)} & 39.4 (2) & 0.79 (1) & 240 (1) \\

    Div. Sum. & \textbf{0.31 (1)} & \textbf{39.5 (1)} & 0.79 (1) & 240 (1) \\

    Rand. Samp. & 0.29 (5) & 37.6 (6) & 0.79 (1) & 240 (1) \\
       
    \bottomrule
  \end{tabular}%
  }
  \caption{Average performance of different LLMs for news categorization on 20 NewsGroup over 5 runs.}
  \label{tbl: 20NewsGroup Summary}
  \vspace{-0.3em}
\end{table}

For the 20 NewsGroup dataset (see \Cref{tbl: 20NewsGroup Summary}) both `Sum.' and `Div. Sum.' approaches achieve the highest F1 scores. With a 39.5\% accuracy, `Div. Sum' outperforms all other scenarios in this metric. Importantly, all approaches achieve better results than providing the full context to the LLM, showing the effectiveness of summarising the information provided to the LLM for this dataset.

\begin{table}[th]
  \smallskip
  \resizebox{0.49\textwidth}{!}{%
  \begin{tabular}{crrrr}
    \toprule
    Scenario & Mac. F1 & Acc.  & Avg. Lat. & Inp. Len.  \\
    \midrule

    Full & 0.51 (6) & 54.0 (6) & 1.07 (6) & 1150 (6) \\

    Full+Sum. & 0.50 (7) & 53.7 (7) & 1.72 (7) & 1400 (7) \\

    First Sent. & \textbf{0.61 (1)} & \textbf{64.5 (1)} & 0.78 (1) & 230 (1) \\

    Last Sent. & 0.53 (4) & 57.9 (4) & 0.78 (1) & 230 (1) \\

    Sum. & 0.56 (3) & 59.8 (3) & 0.78 (1) & 230 (1) \\

    Div. Sum. & 0.58 (2) & 60.9 (2) & 0.78 (1) & 230 (1) \\

    Rand. Samp. & 0.53 (4) & 57.6 (5) & 0.78 (1) & 230 (1) \\
       
    \bottomrule
  \end{tabular}%
  }
  \caption{Average performance of different LLMs for news categorization task on BBC News over 5 runs.}
  \label{tbl: BBC News Summary}
\end{table}

\begin{figure}[h!]
    \centering
    \vspace{-0.5em}
    \begin{subfigure}[b]{0.48\textwidth}
        \centering
        \includegraphics[width=1\textwidth]{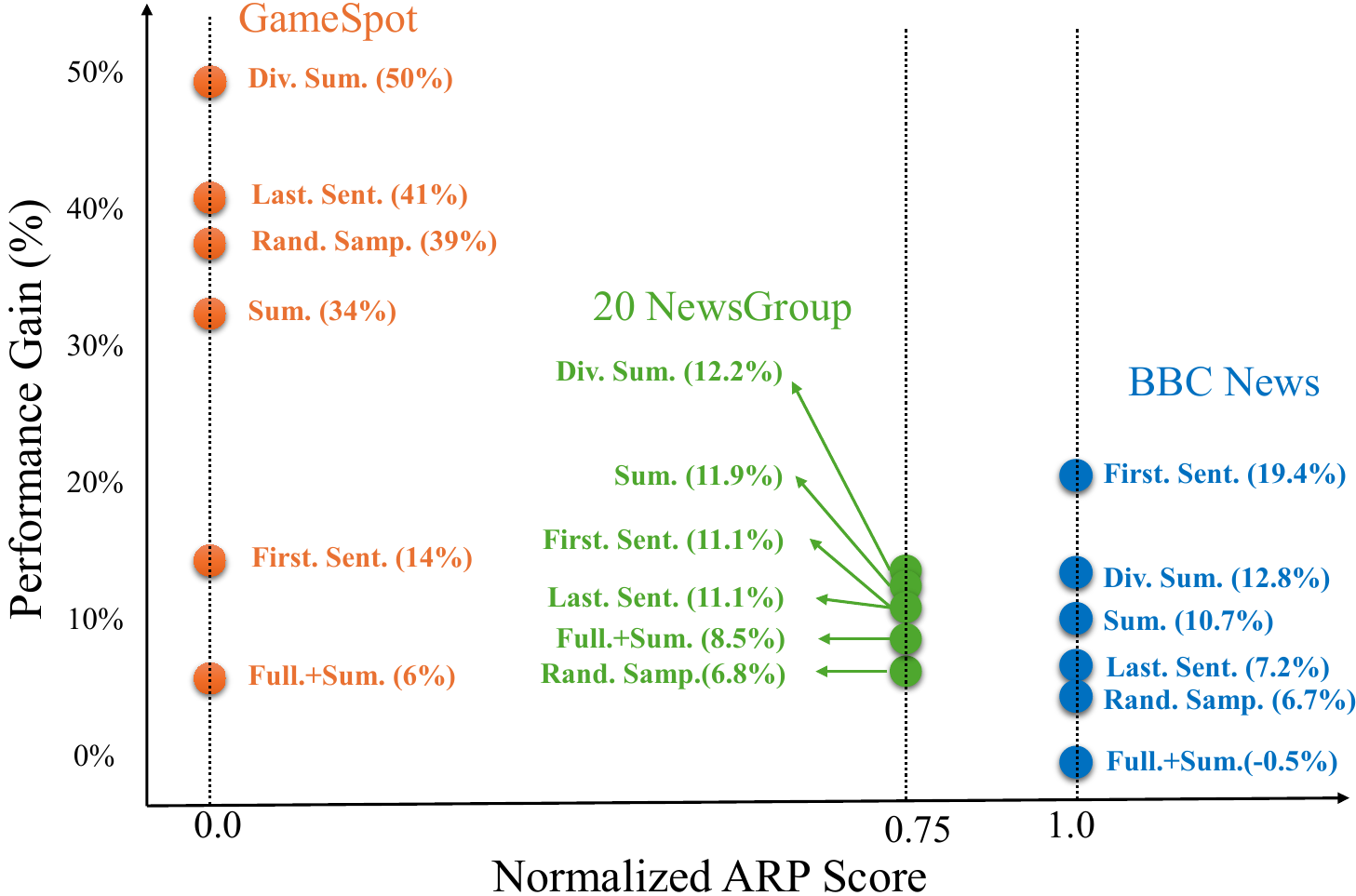}
    \end{subfigure}
    \caption{Performance gain (\% accuracy boost) vs. normalized ARP score for each summarization/truncation scenario compared to the `Full' baseline. Lower ARP scores (more cohesive corpora) generally yield higher performance gains across scenarios.}
    \vspace{-1em}
    \label{fig: ARP vs Performance Gain}
\end{figure}

As summarized in \Cref{tbl: BBC News Summary}, we observe a similar trend for the BBC dataset:  all approaches except `Full+Sum.' outperform `Full' scenario in all metrics. This emphasizes the importance of selectively summarising the information provided to LLMs.

\subsubsection{Ablation Studies}
To further investigate the performance of our models under different conditions, we conducted two ablation studies. First, we varied the truncation/summary lengths from 3 to 15 sentences for all models and evaluated their performance on the GameSpot dataset (\Cref{fig: Claude GameSpot Performance,fig: Avg performance gamespot}). Our solutions consistently outperformed the baseline across different lengths. As you can see from \Cref{fig: Avg performance gamespot}, providing the LLM with as little as 4-5 sentences truncated from the text (whether from its beginning sentences, ending sentences, or even randomly) yields generally better results compared to the LLMs processing the full document. This study aims to isolate the impact of length, revealing LLMs' performance degradation beyond 7-8 sentences. The performance curves for truncation methods (First Sentences and Last Sentences) provide clear evidence of the relationship between input length and model performance across the 3-15 sentence range. We elaborate on these findings in \Cref{par: Length Isolation}.

In the second ablation, we studied the effect of temperature on the Mistral and Llama 3 Instruct models for news categorization on the 20 NewsGroup dataset (\Cref{fig: Temperature Ablation}). Since we are using LLMs for classification and categorization purpose, we try temperatures below 0.1 as higher temperatures lead to higher randomness which is not ideal for classification purposes. We try 0.000, 0.025, 0.050, 0.075, and 0.100 temperatures to see how the performance of the models is affected when changing only this variable. We see that across all these temperatures, the summarization approaches demonstrate superior performance compared to the ``Full Text'' baseline. This experiment demonstrates that LLMs' performance degradation with long sequences persists across different temperature settings, indicating that the observed limitations in handling long texts cannot be attributed to the models' probabilistic sampling behavior.








       


%
\section{Discussion}
\label{sec: Discussions}
\paragraph{Cohesiveness-Performance Gain Study:}We conducted a preliminary study analyzing the correlation between document cohesiveness and performance gains attained by truncation and summarization approaches for each corpus. Using Average Relative Proximity (ARP) score \citep{GhinassiWNP2023} with average cosine similarity scoring over 2-sentence segments, we found that more cohesive corpora (lower ARP) had higher average performance gains from summarization (\Cref{fig: ARP vs Performance Gain}). Future studies could establish these findings further.

\paragraph{Isolation of Length as a Contributing Factor:} \label{par: Length Isolation} This study quantitatively assesses Large Language Models' (LLMs) performance degradation with lengthy inputs, even for straightforward tasks, and proposes summarization as a mitigation strategy. Our findings demonstrate the efficacy of summarization in addressing length-related challenges. However, the performance improvements extend beyond length reduction alone. This is because summarization inherently identifies salient information crucial for tasks like Sentiment Analysis and News Categorization. Our ablation study (see \Cref{fig: Avg performance gamespot}) isolates the length factor by demonstrating that with simple truncation methods such as First Sentences, performance metrics plateau after 7-8 sentences, corroborating our central argument about LLMs' inherent difficulties in processing longer sequences.

\paragraph{Findings and Future Directions:}This study's findings highlight that despite long context windows, SotA LLMs still struggle to effectively process long text sequences, a critical limitation under-examined by prior research for common NLP tasks relying on contextual understanding. Our results highlight the need for more research into optimizing lengthy text inputs to enhance LLM performance. Additionally, we advocate for the development of specialized datasets where input length serves as the primary controlled variable while maintaining consistent content complexity. Such datasets would enable researchers to systematically analyze LLMs' performance degradation with increasing sequence length, particularly for tasks like sentiment analysis and news categorization where contextual understanding is crucial.

\section{Conclusions}
This paper examined the performance of various LLMs (Claude 3 Haiku, Gemini 1 Pro, GPT 3.5 Turbo, Llama 3 Instruct, and Mistral Instruct) on lengthy inputs in core NLP tasks. The results, across three datasets and two tasks, consistently indicate longer inputs result in worse performance. Further ablation experiments on truncation/summarization length and model temperature disqualified the random sampling behaviour of the LLMs to be a confounding factor and showcased LLMs struggles in even handling moderately lengthy inputs. We proposed several ad-hoc solutions to substantially enhance LLMs' performance (up to 50\%) on long input sequences and reduce API cost (up to 93\%) and average latency (up to 50\%).

\section*{Acknowledgements}
This work was partially supported by  UKRI via the Centre for Doctoral Training in Intelligent Games and Game Intelligence (IGGI, EP/S022325/1) and projects AP4L (EP/W032473/1), ARCIDUCA (EP/W001632/1) and AdSoLve (Responsible AI UK, EP/Y009800/1, project KP0016); and by the Slovenian Research Agency via research core funding for the programme Knowledge Technologies (P2-0103) and the project EMMA (L2-50070).

We also thank Mehran Hosseini and Zahraa Al-Sahili for their comments and discussions which helped us improve this manuscript.

\section*{Limitations}
\paragraph{Rapid change of model specifications.} First, we examined a diverse set of SotA large language models, all of which are among the most commonly used LLMs. However, the rapidly evolving nature of this field means these findings may not be fully generalized to future LLMs with different architectures and training paradigms.

\paragraph{Tasks requiring general context understanding.} Second, although we evaluated performance on two core NLP tasks (sentiment analysis and news categorization) across three datasets, further research is needed to determine if our conclusions hold for a wider range of natural language understanding tasks and domains. The datasets we used, while lengthy, may not fully capture the types of long-form content that LLMs will need to process in other real-world applications. However, our studies here laid the foundation to show the limitations of LLMs when dealing with long sequences even in canonical NLP tasks as an underexplored problem. We believe our ad-hoc solutions are applicable to a wide variety of tasks that require a general understanding of the input sequence rather than a detailed understanding of the whole context.

\paragraph{Societal Impact and Ethical Considerations} In terms of societal impact, we believe our findings can help enable the more effective and efficient application of LLMs to tasks involving longer documents, which has the potential to unlock significant value in domains like business analytics, legal contract review, and scientific literature mining. In addition, our ad-hoc solutions are a step forward for democratizing access to AI. Even though using LLM APIs is becoming more affordable, our approaches reduce the API cost for users by up to 93\% and this further enables more accessibility across different sections of society. At the same time, the ability to extract key information from lengthy privacy policies, terms of service, and other consumer agreements could be misused in ways that fail to represent the full context. As LLMs achieve greater summarization capabilities, extra care will be needed to ensure these summaries are accurate, unbiased and not misused for deceptive purposes.

Overall, while our work provides important empirical insights into the limitations of current LLMs on long sequence tasks and highlights promising directions for overcoming these challenges, we see it as a motivating starting point rather than a conclusive result. We encourage the research community to further test and expand on our findings to drive the development of more capable and robust prompting techniques.

\bibliography{anthology,custom}

\newpage

\appendix
\label{sec:appendix}
\section{Appendix}
\label{sec:appendix}

\subsection{Datasets and Main Experiments}
We used three datasets in our evaluations. Here, we provide a more detailed explanation of each dataset and task.

\subsubsection{Sentiment Analysis}
\paragraph{GameSpot.} The GameSpot Reviews dataset contains over 12,000 lengthy video game reviews with author-assigned sentiment scores ranging from 1 to 100. Almost all the reviews in this dataset are quite lengthy and by using a minimum threshold of 45 sentences, there are still thousands of reviews available in this dataset.

In the sentiment analysis experiments, we asked each LLM to give a sentiment rating of 1 to 100 for each document. Most labels in the data were multiples of 10 (i.e., 10, 20, 30, \ldots, 90, 100). However, sometimes the labels had other values like 95, or 85 as well. To this end and to cover even the corner cases, we asked the LLM to predict the label as an integer from 1 to 100. The calculation of MSE and MAE metrics is straightforward and according to the standard definition. For calculating the accuracy, we considered a prediction as accurate if it was within 5 scoring points of the label. For example, if the label has a value of 70, a predicted label between 65-75 range is considered an accurate prediction and any prediction outside this range is considered not accurate.

Regarding the temperature used in this study as well as other studies, we tried different temperature values for the LLMs but no significant change or decrease was observed by doing this and summarization/truncation methods always showed superior performance. The results reported in this experiment are the average over 5 runs and where applicable we have reported the 85\% Confidence Interval as well. The temperature in the experiments reported here was set to 0.01. 

\subsection{News Categorization}
\paragraph{20 NewsGroup.} The 20 Newsgroups dataset features nearly 20,000 documents across 20 topic categories like politics, religion, sports and computers. We focused on the subset longer than 60 sentences, averaging 3450 tokens per document when tokenized by the NLTK word tokenizer. 

\paragraph{BBC News Archive.} The BBC News Archive, consisting of 2,225 articles covering business, entertainment, politics, sports and tech. We focused our study on the subset with more than 35 sentences, averaging 1150 tokens per document when tokenized by the NLTK word tokenizer.

\begin{figure*}[ht]
    \centering
    \begin{subfigure}[b]{\textwidth}
        \centering
        \includegraphics[width=\textwidth]{figs/length_analysis/legend.pdf}
    \end{subfigure}

    \begin{subfigure}[b]{0.48\textwidth}
        \centering
        \includegraphics[width=1.05\textwidth]{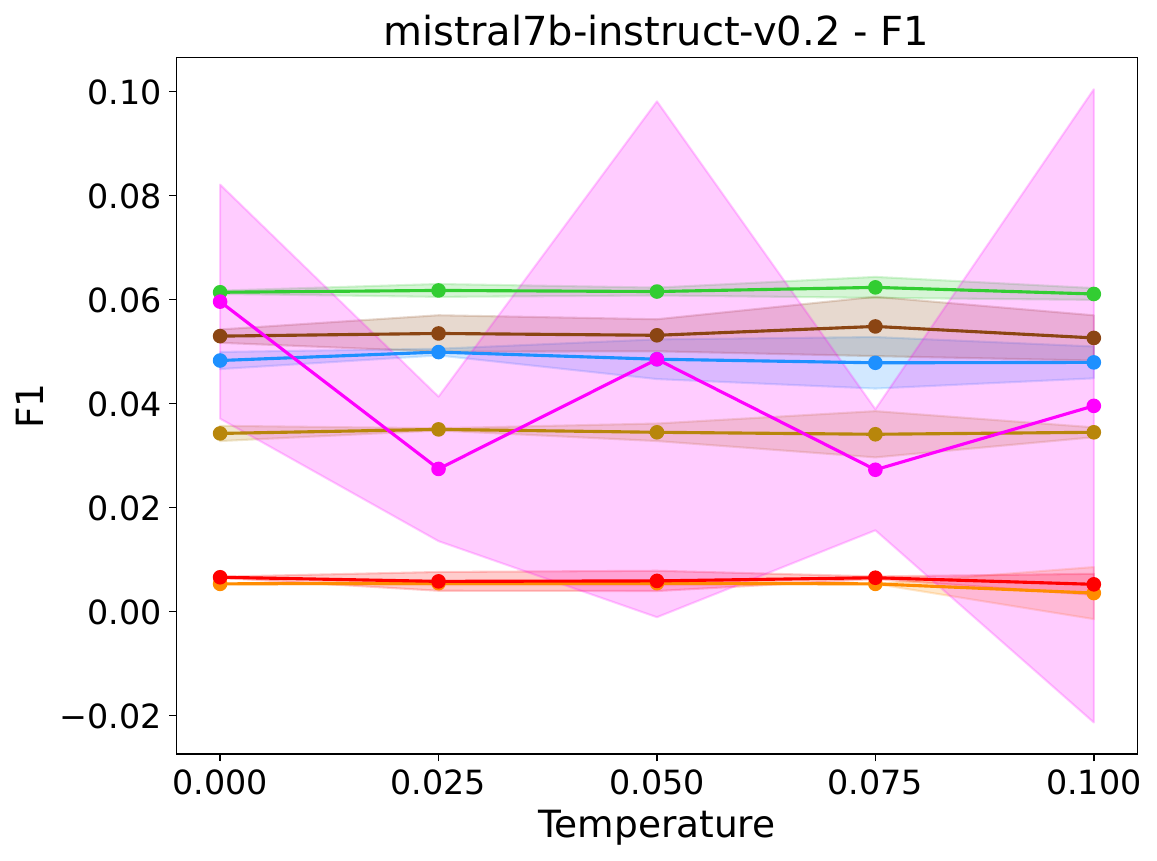}
        \vspace{-1.6em}
        \caption{Mistral 7b Instruct F1-Temperature Curve}
        \vspace{1em}
    \end{subfigure}
    \hfill
    \begin{subfigure}[b]{0.48\textwidth}
        \centering
        \includegraphics[width=1.05\textwidth]{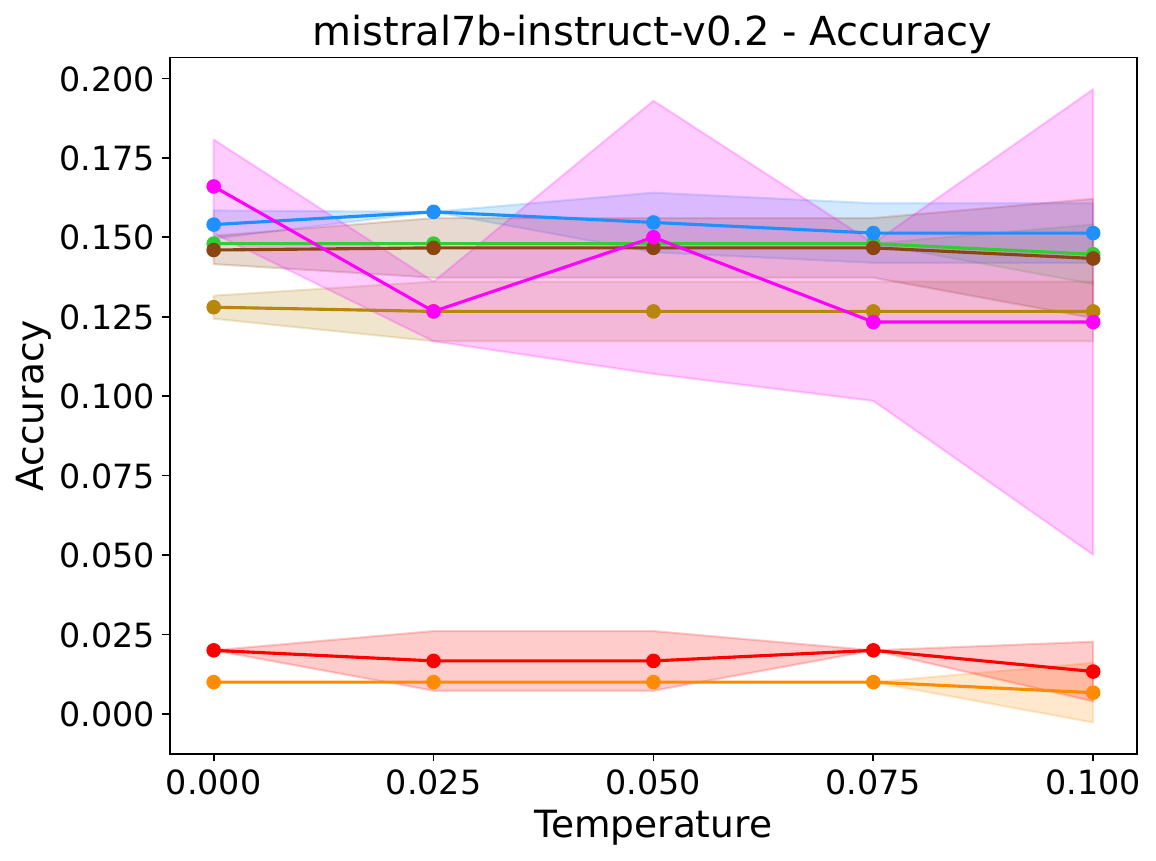}
        \vspace{-1.6em}
        \caption{Mistral 7b Instruct Accuracy-Temperature Curve}
        \label{fig: Avg MAE mistral7b-gamespot}
        \vspace{1em}
    \end{subfigure}

    \begin{subfigure}[b]{0.48\textwidth}
        \centering
        \includegraphics[width=1.05\textwidth]{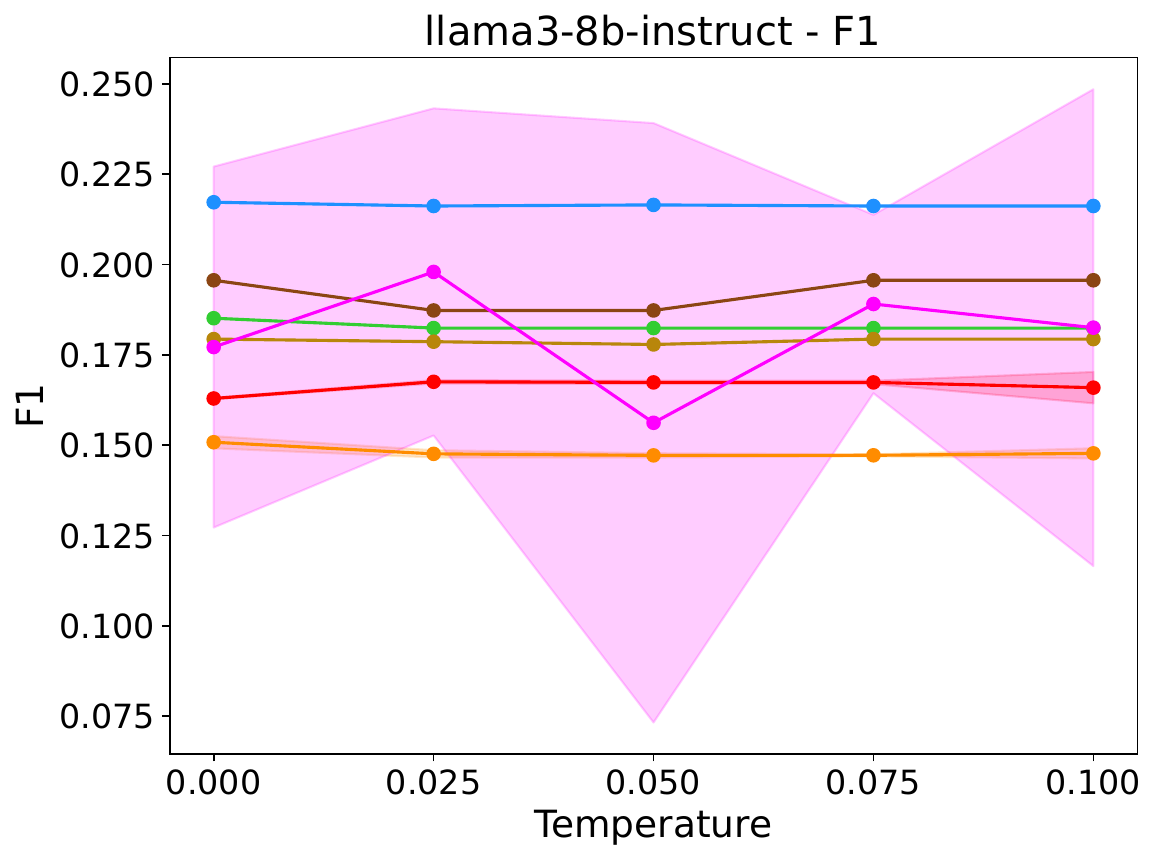}
        \vspace{-1.6em}
        \caption{Llama3 8b Instruct F1-Temperature Curve}
        \label{fig: Avg Acc llama3-gamespot}
        \vspace{1em}
    \end{subfigure}
    \hfill
    \begin{subfigure}[b]{0.48\textwidth}
        \centering
        \includegraphics[width=1.05\textwidth]{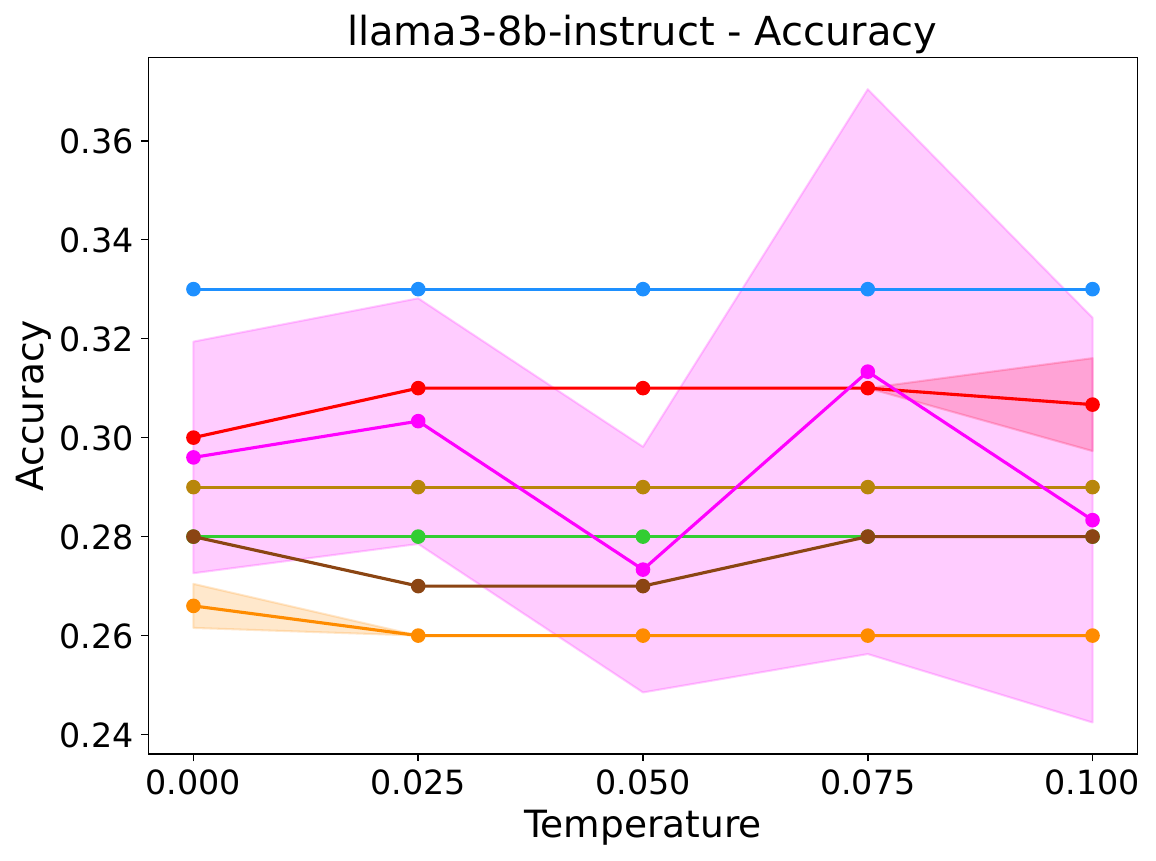}
        \vspace{-1.6em}
        \caption{Llama3 8b Instruct Accuracy-Temperature Curve}
        \label{fig: Avg MAE llama3-gamespot}
        \vspace{1em}
    \end{subfigure}
    \vspace{-1em}
    \caption{Ablation Study on temperature in the news categorization task on 20 newsgroup dataset. The results show the performance of the models does not experience much difference as we change the temperature from 0 to 0.1 aside from a slight increase in the confidence interval. These results are over 5 runs. The width of the 85\% Confidence Interval for the `Random Sampling' scenario is much bigger due to the randomness introduced by selecting the sentences. }
    \label{fig: Temperature Ablation}
\end{figure*}

As we explain in the main body of the paper as well as the results shown in \cref{fig: Temperature Ablation}, when trying different temperatures, there were no meaningful changes in the results and order of the approaches in terms of their performance.

\begin{table}[th]
  \smallskip
  \resizebox{0.49\textwidth}{!}{%
  \begin{tabular}{clrrrrr}
    \toprule
    Model & Scenario & Mac. F1 & Acc.  & Avg. Lat. & Inp. Len.  \\
    \midrule

    & Full & 0.54 & 67.8 & 1.24 & 3450  \\

    & Full+Sum. & 0.58 & 71.2 & 1.78 & 3700 \\

    & First Sent. & 0.50 & 66.4 & 0.72 & 240\\

    & Last Sent. & 0.49 & 64.4 & 0.72 & 240 \\

    & Sum. & 0.56 & 69.6 & 0.72 & 240 \\

    & Div. Sum. & 0.52 & 65.5  & 0.72 & 240 \\

    \multirow{-7}{*}{\rotatebox[origin=c]{90}{Claude 3 Haiku}} & Rand. Samp. & 0.50 & 64.6 & 0.72 & 240 \\
       
    \midrule

    & Full & 0.38 & 47.1 & 1.19 & 3450 \\

    & Full+Sum. & 0.41 & 53.2 & 1.95 & 3700 \\

    & First Sent. & 0.42 & 45.2 & 0.33 & 240 \\

    & Last Sent. & 0.40  & 42.8 & 0.33 & 240 \\

    & Sum. & 0.38 & 43.4 & 0.33 & 240 \\

    & Div. Sum. & 0.39 & 44.3 & 0.33 & 240 \\

    \multirow{-7}{*}{\rotatebox[origin=c]{90}{GPT 3.5 Turbo}} & Rand. Samp. & 0.41 & 43.1 & 0.33 & 240 \\
       
    \midrule

    & Full & 0.30 & 34.6 & 3.48 & 3450 \\

    & Full+Sum. & 0.36 & 35.2 & 3.88 & 3700\\

    & First Sent. & 0.36 & 43.2 & 1.12 & 240 \\

    & Last Sent. & 0.45 & 46.4 & 1.12 & 240 \\

    & Sum. & 0.39 & 40.4 & 1.12 & 240 \\

    & Div. Sum. & 0.38 & 40.8 & 1.12 & 240 \\

    \multirow{-7}{*}{\rotatebox[origin=c]{90}{Gemini Pro}} & Rand. Samp. & 0.36 & 40.2 & 1.12 & 240 \\
       
    \midrule

    & Full & 0.01 & 1.1 & 0.96 & 3450 \\

    & Full+Sum. & 0.01 & 2.3 & 0.99 & 3650 \\

    & First Sent. & 0.03 & 11.5 & 0.87 & 180 \\

    & Last Sent. & 0.04 & 12.2 & 0.87 & 180 \\

    & Sum. & 0.05 & 12.3 & 0.87 & 180 \\

    & Div. Sum. & 0.04 & 12.6 & 0.87 & 180\\

    \multirow{-7}{*}{\rotatebox[origin=c]{90}{Mistral 7b}} & Rand. Samp. & 0.05 & 12.2 & 0.87 & 180 \\
       
    \midrule

    & Full & 0.15 & 25.6 & 1.02 & 3450 \\

    & Full+Sum. & 0.16 & 29.3 & 1.11 & 3650 \\

    & First Sent. & 0.17 & 29.1 & 0.89 & 180 \\

    & Last Sent. & 0.17 & 29.8 & 0.89 & 180 \\

    & Sum. & 0.19 & 31.2 & 0.89 & 180 \\

    & Div. Sum. & 0.21 & 34.1 & 0.89 & 180 \\

    \multirow{-7}{*}{\rotatebox[origin=c]{90}{Llama 3 8b}} & Rand. Samp. & 0.16 & 27.8 & 0.89 & 180 \\
    
    \bottomrule
  \end{tabular}%
  }
  \caption{The performance of different LLMs for Categorization task on 20 NewsGroup dataset. $N$ parameter for summarization and truncation is set to 7 for the Mistral and Llama models and 10 for the others.}
  \label{tbl: 20NewsGroup}
\end{table}
\begin{table}[th]
  \smallskip
  \resizebox{0.49\textwidth}{!}{%
  \begin{tabular}{clrrrrr}
    \toprule
    Model & Scenario & Mac. F1 & Acc.  & Avg. Lat. & Inp. Len.  \\
    \midrule

    & Full & 0.63 & 63.8 & 0.69 & 1150 \\

    & Full+Sum. & 0.67 & 67.1 & 1.45 & 1400 \\

    & First Sent. & 0.69 & 70.4 & 0.65 & 230 \\

    & Last Sent. & 0.56 & 56.9 & 0.65 & 230 \\

    & Sum. & 0.61 & 61.5 & 0.65 & 230 \\

    & Div. Sum. & 0.64 & 63.8  & 0.65 & 230 \\

    \multirow{-7}{*}{\rotatebox[origin=c]{90}{Claude 3 Haiku}} & Rand. Samp. & 0.60 & 61.2 & 0.65 & 230 \\
       
    \midrule

    & Full & 0.75 & 83.8 & 0.54 & 1150 \\

    & Full+Sum. & 0.67 & 76.4 & 1.08 & 1400 \\

    & First Sent. & 0.80 & 86.3 & 0.49 & 230 \\

    & Last Sent. & 0.69 & 78.4 & 0.49 & 230 \\

    & Sum. & 0.72 & 79.9 & 0.49 & 230 \\

    & Div. Sum. & 0.69 & 78.7 & 0.49 & 230 \\

    \multirow{-7}{*}{\rotatebox[origin=c]{90}{GPT 3.5 Turbo}} & Rand. Samp. & 0.68 & 77.7 & 0.49 & 230 \\
       
    \midrule

    & Full & 0.65 & 61.2 & 2.26 & 1150 \\

    & Full+Sum. & 0.69 & 66.6 & 2.61 & 1400 \\

    & First Sent. & 0.63 & 59.7 & 1.04 & 230 \\

    & Last Sent. & 0.64 & 58.4 & 1.04 & 230 \\

    & Sum. & 0.61 & 57.7 & 1.04 & 230 \\

    & Div. Sum. & 0.61 & 56.8 & 1.04 & 230 \\

    \multirow{-7}{*}{\rotatebox[origin=c]{90}{Gemini Pro}} & Rand. Samp. & 0.61 & 56.2  & 1.04 & 230 \\
       
    \midrule

    & Full & 0.13 & 22.1 & 0.97 & 1150 \\

    & Full+Sum. & 0.03 & 11.7 & 1.67 & 1330 \\

    & First Sent. & 0.32 & 37.1 & 0.85 & 170 \\

    & Last Sent. & 0.23 & 32.1 & 0.85 & 170 \\

    & Sum. & 0.33 & 38.5 & 0.85 & 170 \\

    & Div. Sum. & 0.35 & 39.9 & 0.85 & 170 \\

    \multirow{-7}{*}{\rotatebox[origin=c]{90}{Mistral 7b}} & Rand. Samp. & 0.28 & 33.5 & 0.85 & 170 \\
       
    \midrule

    & Full & 0.38 & 39.2 & 0.89 & 1150 \\

    & Full+Sum. & 0.42 & 46.6 & 1.81 & 1330 \\

    & First Sent. & 0.62 & 68.8 & 0.85 & 170 \\

    & Last Sent. & 0.53 & 63.5 & 0.85 & 170 \\

    & Sum. & 0.54 & 61.2 & 0.85 & 170 \\

    & Div. Sum. & 0.59 & 65.3 & 0.85 & 170 \\

    \multirow{-7}{*}{\rotatebox[origin=c]{90}{Llama 3 8b}} & Rand. Samp. & 0.51 & 59.2 & 0.85 & 170 \\
    
    \bottomrule
  \end{tabular}%
  }
  \caption{The performance of different LLMs for Categorization task on BBC News dataset. $N$ parameter for summarization and truncation is set to 7 for the Mistral and Llama models and 10 for the others.}
  \label{tbl: BBC News}
\end{table}

The results reported in this paper in the tables for the news categorization task are averaged over 5 runs and we have reported 85\% Confidence Interval when applicable. The temperature of the models was set to 0.0 for the results reported in the tables.

\subsubsection{Diverse Summarization}
\label{app: diverse summarization}
Here we provide more explanation about the diverse summarization approach and how the green-coloured component (in \Cref{fig: Diverse Summary Schemes}) which is the diversity selector in our algorithm works.

To write equations describing what the green component is doing, we can focus on the main functions in this component and their inputs and outputs. Let's denote the input text as $T$, the number of desired sentences as $N$, and the set of sentences in the text as $S = {s_1, s_2, \ldots, s_{M}}$.

\begin{enumerate}[label=\arabic*.]
\item Tokenize Sentences:
\begin{equation}
S = \mathrm{sent\_tokenize}(T)
\end{equation}

\item Calculate sentence embeddings:
\begin{equation}
E = \mathrm{TfidfVectorizer}(S)
\end{equation}
where $E$ is the TF-IDF matrix representing the embeddings of the sentences.

\item Calculate diversity scores:
\begin{align}
D &= 1 - \mathrm{cos\_sim}(E) \label{eq: calculate cosine sim}\\ 
D_{\mathrm{sum}} &= \sum_{i=1}^{M} D_i \label{eq: sum diversity}
\end{align}
where $D$ is the dissimilarity matrix, and $D_{\mathrm{sum}}$ is the sum of dissimilarity scores for each sentence.

\item Select top N diverse sentences:
\begin{equation}
S_{\mathrm{top_N}} = \arg\max_N(D_{\mathrm{sum}})
\end{equation}
where $S_{\mathrm{top_N}}$ is the set of $N$ sentences with the highest diversity scores.

\item Generate the final summary by joining the sentences:
\begin{equation}
\mathrm{summary} = \mathrm{join}(S_{\mathrm{top_N}})
\end{equation}
where the final summary is the concatenation of the selected top $N$ diverse sentences.

\end{enumerate}

The main steps denoted in the above equations can be summarized as follows:
\begin{enumerate}
\item Tokenize the input text $T$ into a set of sentences $S$.
\item Compute the TF-IDF embedding matrix $E$ for the sentences.
\item Calculate the dissimilarity matrix $D$ using cosine similarity and sum the dissimilarity scores for each sentence. In \Cref{eq: calculate cosine sim}, $D$ represents the dissimilarity matrix, which is calculated as 1 minus the cosine similarity of the sentence embeddings $E$. This means $D$ contains the pairwise dissimilarity scores between all sentences. In \Cref{eq: sum diversity}, $D\_sum$ is calculated by summing the dissimilarity scores for each sentence. This means $D\_sum$ is a vector where each element represents the total dissimilarity score for a particular sentence compared to all other sentences.
\item Select the top $N$ sentences $S_{\mathrm{top_N}}$ with the highest diversity scores.
\item Concatenate the selected sentences to form the final summary.
\end{enumerate}

\begin{table*}[th]
  \centering
  \smallskip
  \resizebox{0.8\textwidth}{!}{%
  \begin{tabular}{clrrrrr}
    \toprule
    Model & Scenario & MSE & MAE & Accuracy & Avg. Lat. & Inp. Len. \\
    \midrule

    & Full & 112.3 (6) & 8.50 (6) & 43.8 (7) & 1.04 & 2120 \\

    & Full+Sum. & 80.2 (2) & 7.17 (2) & 50.6 (5) & 1.05 & 2450 \\

    & First Sent. & 131.8 (7) & 8.87 (7)  & 44.4 (6) & 0.77 & 320 \\

    & Last Sent. & \textbf{77.0 (1)} & \textbf{6.58 (1)} & 60.2 (2) & 0.77 & 320 \\

    & Sum. & 97.6 (3) & 7.55 (4) & 56.8 (4) & 0.77 & 320 \\

    & Div. Sum. & 106.0 (5) & 7.21 (3) & \textbf{65.2 (1)} & 0.77 & 320 \\

    \multirow{-7}{*}{\rotatebox[origin=c]{90}{Claude 3 Haiku}} & Rand. Samp. & 104.2 (4) & 7.56 (5) & 59.0 (3) & 0.77 & 320 \\
       
    \midrule

    & Full & 134.9 (4) & 9.76 (6) & 40.4 (7) & 0.59 & 2120  \\

    & Full+Sum. & \textbf{110.8 (1)} & \textbf{8.75 (1)} & 46.6 (5) & 0.61 & 2450 \\

    & First Sent. & 177.3 (7) & 10.81 (7) & 41.0 (6) & 0.47 & 320 \\

    & Last Sent. & 119.1 (3) & 8.78 (2) & 55.6 (2) & 0.47  & 320 \\

    & Sum. & 117.8 (2) & 8.94 (3) & 53.0 (3) & 0.47 & 320 \\

    & Div. Sum. & 141.7 (6) & 9.11 (4) & \textbf{59.2 (1)} & 0.47 & 320 \\

    \multirow{-7}{*}{\rotatebox[origin=c]{90}{GPT 3.5 Turbo}} & Rand. Samp. & 135.3 (5) & 9.27 (5) & 51.8 (4) & 0.47 & 320 \\
       
    \midrule

    & Full & 130.8 (6) & 9.89 (7) & 43.8 (7) & 2.74 & 2120 \\

    & Full+Sum. & 117.6 (6) & 9.14 (6) & 53.4 (6) & 2.96 & 2450 \\

    & First Sent. & 88.0 (4) & 7.15 (4) & 61.6 (5) & 1.15  & 320 \\

    & Last Sent. & 83.3 (3) & 7.14 (3) & 63.8 (4) & 1.08  & 320 \\

    & Sum. & \textbf{73.3 (1)} & \textbf{6.80 (1)} & 67.4 (2) & 1.08 & 320 \\

    & Div. Sum. & 94.2 (5) & 7.23 (5) & \textbf{69.6 (1)} & 1.08 & 320 \\

    \multirow{-7}{*}{\rotatebox[origin=c]{90}{Gemini Pro}} & Rand. Samp. & 78.8 (2) & 7.05 (2) & 66.2 (3) & 1.08 & 320 \\

    \midrule

    & Full & 811.6 (6) & 19.54 (6) & 13.1 (6) & 1.01 & 2120 \\

    & Full+Sum. & 1532.1 (7) & 30.68 (7) & 8.9 (7) & 1.04 & 2450 \\

    & First Sent. & 236.4 (4) & 10.81 (5) &  29.0 (4) & 0.91 & 320 \\

    & Last Sent. & \textbf{93.5 (1)} & \textbf{7.96 (1)} & \textbf{38.0 (1)} & 0.91 & 320 \\

    & Sum. & 172.5 (4) & 10.86 (4) & 23.1 (5) & 0.91 & 320 \\

    & Div. Sum. & 155.1 (2) & 9.56 (2) & 35.8 (2) & 0.91 & 320 \\

    \multirow{-5}{*}{\rotatebox[origin=c]{90}{Mistral 7b}} & Rand. Samp. & 164.1 (3) & 9.57 (3) & 34.6 (3) & 0.91 & 320 \\

    \midrule

    & Full & 174.7 (5) & 10.62 (5) & 39.1 (3) & 0.95 & 2120 \\

    & Full+Sum. & 192.4 (6) & 11.79 (7) & 31.3 (6) & 0.99 & 2450 \\

    & First Sent. & 212.1 (7) & 11.76 (6) & 30.1 (7) & 0.90 & 320 \\

    & Last Sent. & \textbf{125.4 (1)} & \textbf{9.33 (1)} & 36.1 (5) & 0.90 & 320 \\

    & Sum. & 159.7 (2) & 10.06 (2) & \textbf{40.5 (1)} & 0.90 & 320 \\

    & Div. Sum. & 171.6 (4) & 10.15 (4) & 40.0 (2) & 0.90 & 320 \\

    \multirow{-5}{*}{\rotatebox[origin=c]{90}{Llama 3 8b}} & Rand. Samp. & 165.5 (3) & 10.10  (3) & 38.2 (4) & 0.90 & 320 \\
    
    \bottomrule
  \end{tabular}%
  }
  \caption{The performance of different LLMs for Sentiment Analysis task on GameSpot dataset. $N$ parameter for summarization and truncation is set to 7 for all the models.}
  \label{tbl: GameSpot}
\end{table*}
\begin{figure*}[ht]
    \centering
    \begin{subfigure}[b]{\textwidth}
        \centering
        \includegraphics[width=\textwidth]{figs/length_analysis/legend.pdf}
    \end{subfigure}

    \vspace{-0.3em}

    \begin{subfigure}[b]{0.328\textwidth}
        \centering
        \includegraphics[width=1.05\textwidth]{figs/length_analysis/claude3-haiku/Accuracy_vs_num_sentences.pdf}
        \vspace{-1.75em}
        \caption{Claude3 Haiku Accuracy Curve}
        \label{fig: Avg Acc claude-gamespot}
        \vspace{1em}
    \end{subfigure}
    \hfill
    \begin{subfigure}[b]{0.328\textwidth}
        \centering
        \includegraphics[width=1.05\textwidth]{figs/length_analysis/claude3-haiku/MAE_vs_num_sentences.pdf}
        \vspace{-1.75em}
        \caption{Claude3 Haiku MAE Curve}
        \label{fig: Avg MAE claude-gamespot}
        \vspace{1em}
    \end{subfigure}
    \hfill
    \begin{subfigure}[b]{0.328\textwidth}
        \centering
        \includegraphics[width=1.05\textwidth]{figs/length_analysis/claude3-haiku/MSE_vs_num_sentences.pdf}
        \vspace{-1.75em}
        \caption{Claude3 Haiku MSE Curve}
        \label{fig: Avg MSE claude-gamespot}
        \vspace{1em}
    \end{subfigure}

    \vspace{-0.4em}

    \begin{subfigure}[b]{0.328\textwidth}
        \centering
        \includegraphics[width=1.05\textwidth]{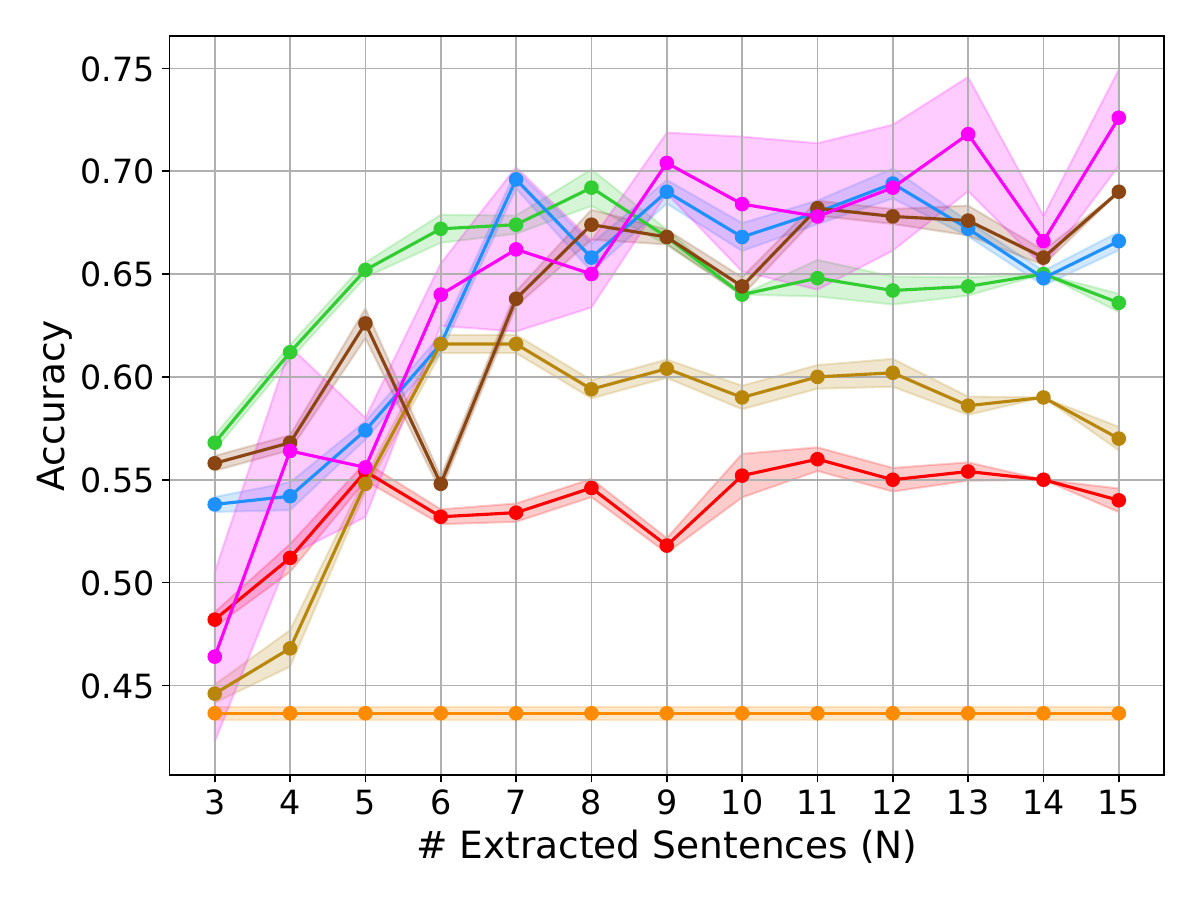}
        \vspace{-1.75em}
        \caption{Gemini Pro Accuracy Curve}
        \label{fig: Avg Acc geminipro-gamespot}
        \vspace{1em}
    \end{subfigure}
    \hfill
    \begin{subfigure}[b]{0.328\textwidth}
        \centering
        \includegraphics[width=1.05\textwidth]{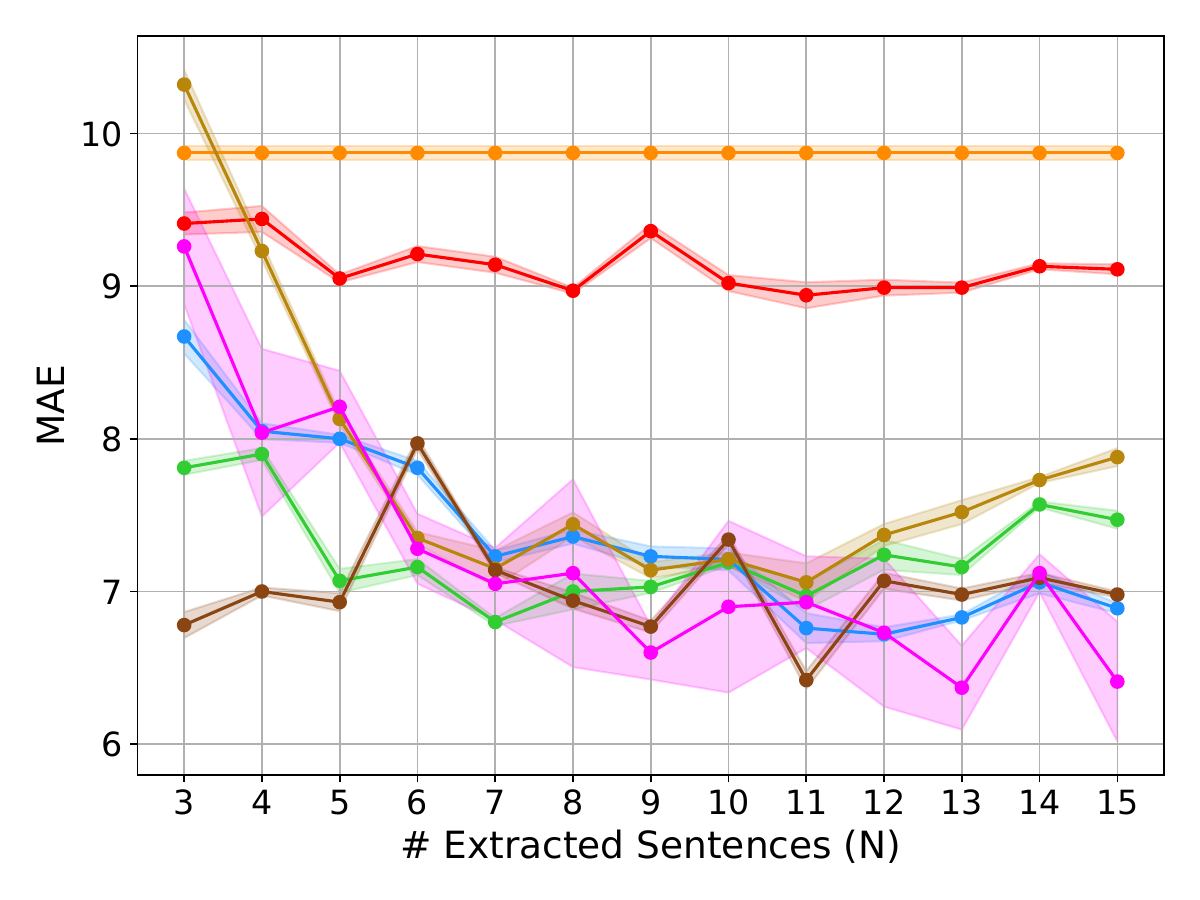}
        \vspace{-1.75em}
        \caption{Gemini Pro MAE Curve}
        \label{fig: Avg MAE geminipro-gamespot}
        \vspace{1em}
    \end{subfigure}
    \hfill
    \begin{subfigure}[b]{0.328\textwidth}
        \centering
        \includegraphics[width=1.05\textwidth]{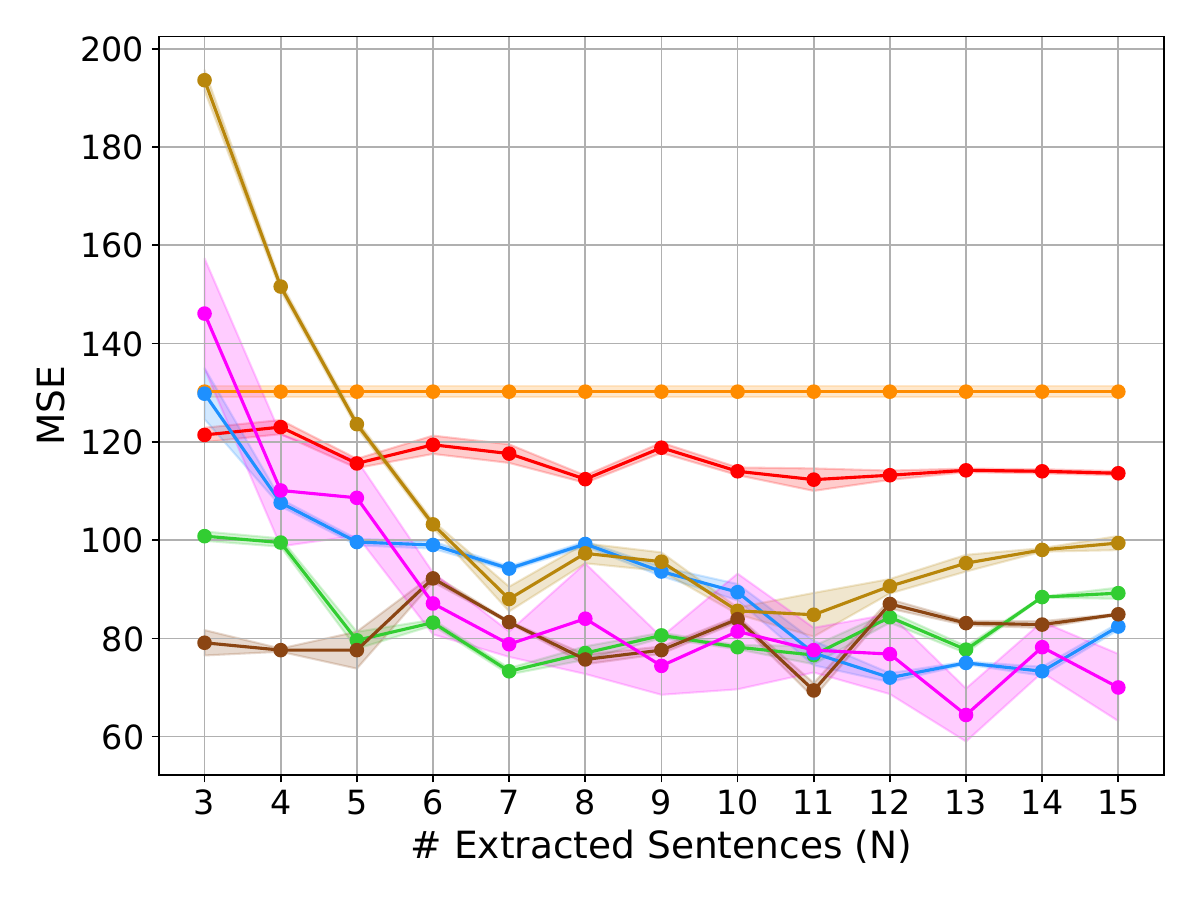}
        \vspace{-1.75em}
        \caption{Gemini Pro MSE Curve}
        \label{fig: Avg MSE geminipro-gamespot}
        \vspace{1em}
    \end{subfigure}

    \vspace{-0.4em}

    \begin{subfigure}[b]{0.328\textwidth}
        \centering
        \includegraphics[width=1.05\textwidth]{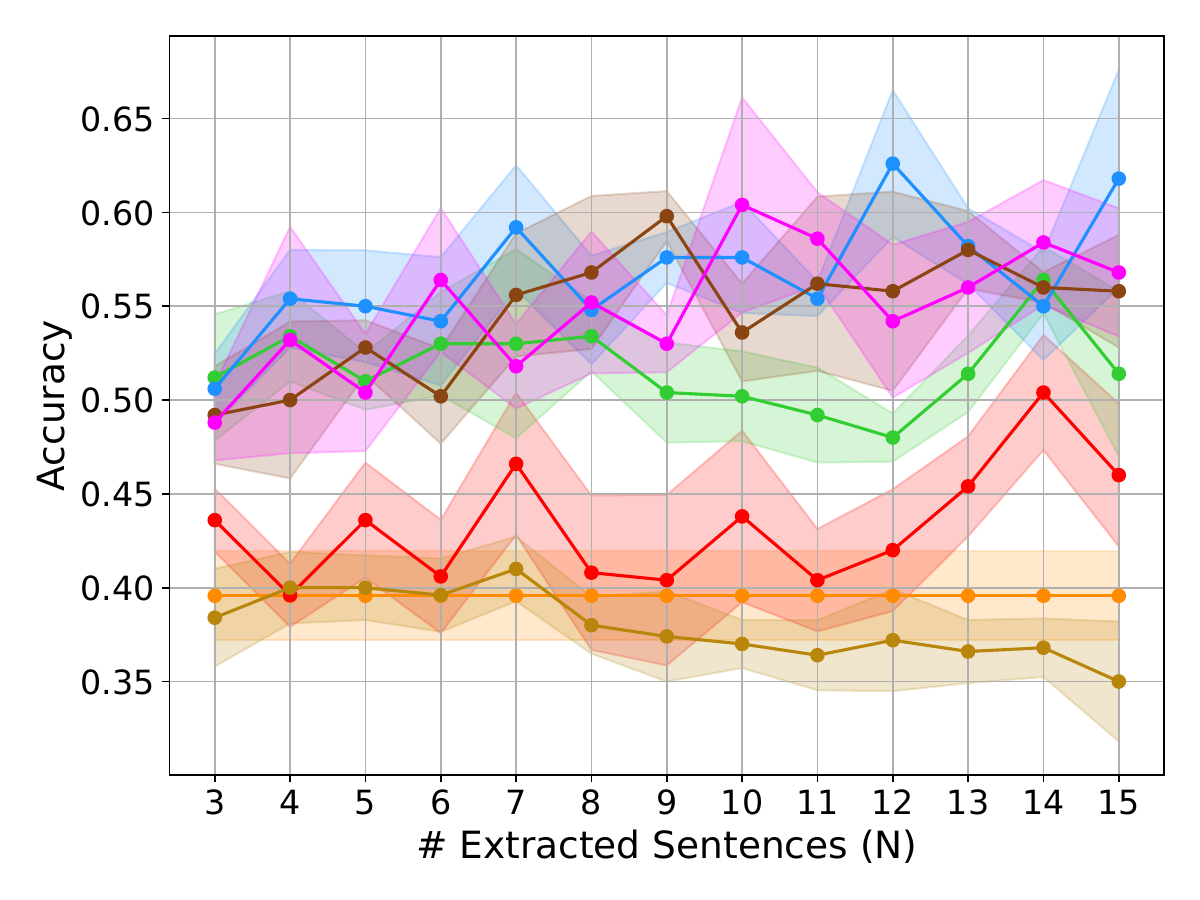}
        \vspace{-1.75em}
        \caption{GPT3.5 Turbo Accuracy Curve}
        \label{fig: Avg Acc gpt3.5-gamespot}
        \vspace{1em}
    \end{subfigure}
    \hfill
    \begin{subfigure}[b]{0.328\textwidth}
        \centering
        \includegraphics[width=1.05\textwidth]{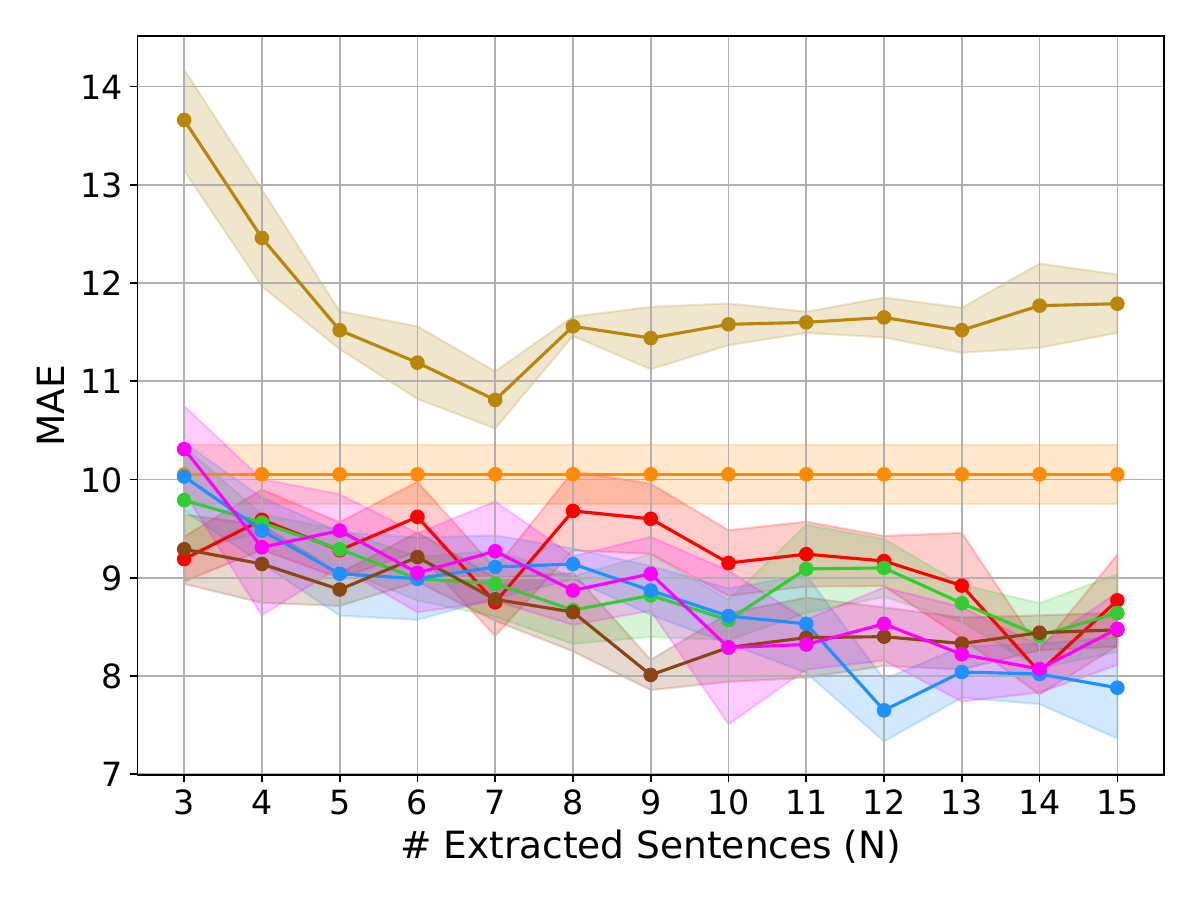}
        \vspace{-1.75em}
        \caption{GPT3.5 Turbo MAE Curve}
        \label{fig: Avg MAE gpt3.5-gamespot}
        \vspace{1em}
    \end{subfigure}
    \hfill
    \begin{subfigure}[b]{0.328\textwidth}
        \centering
        \includegraphics[width=1.05\textwidth]{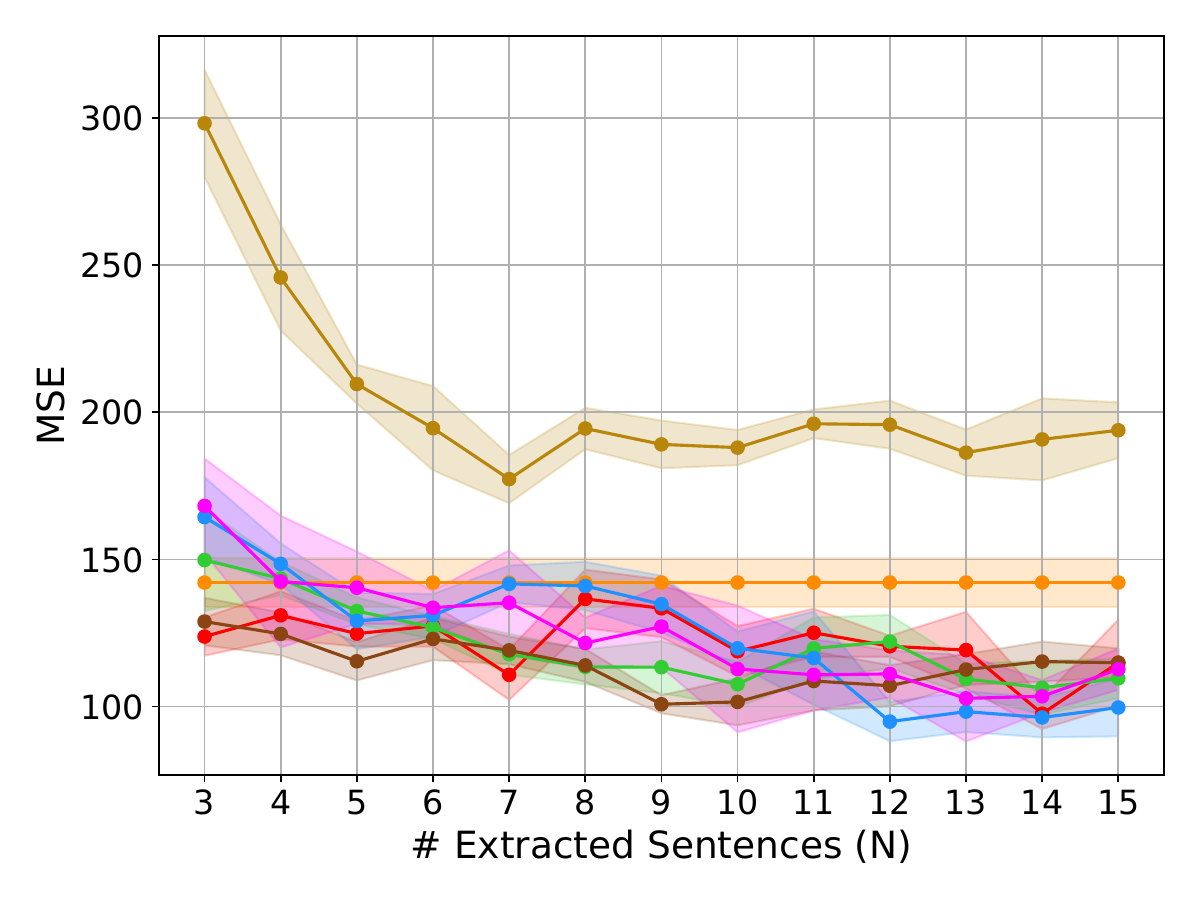}
        \vspace{-1.75em}
        \caption{GPT3.5 Turbo MSE Curve}
        \label{fig: Avg MSE gpt3.5-gamespot}
        \vspace{1em}
    \end{subfigure}

    \vspace{-0.4em}

    \begin{subfigure}[b]{0.328\textwidth}
        \centering
        \includegraphics[width=1.05\textwidth]{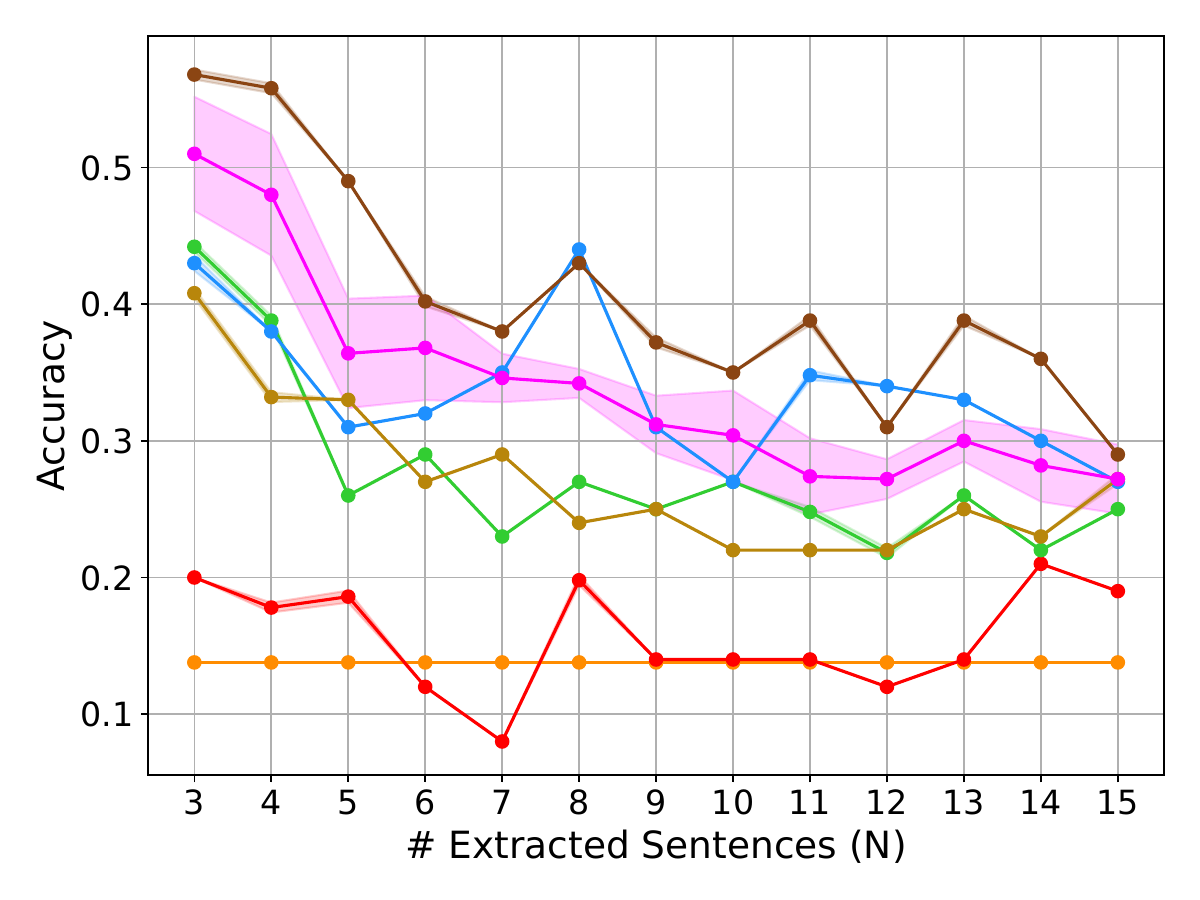}
        \vspace{-1.75em}
        \caption{Mistral 7b Instruct Accuracy Curve}
        \label{fig: Avg Acc mistral7b-gamespot}
        \vspace{1em}
    \end{subfigure}
    \hfill
    \begin{subfigure}[b]{0.328\textwidth}
        \centering
        \includegraphics[width=1.05\textwidth]{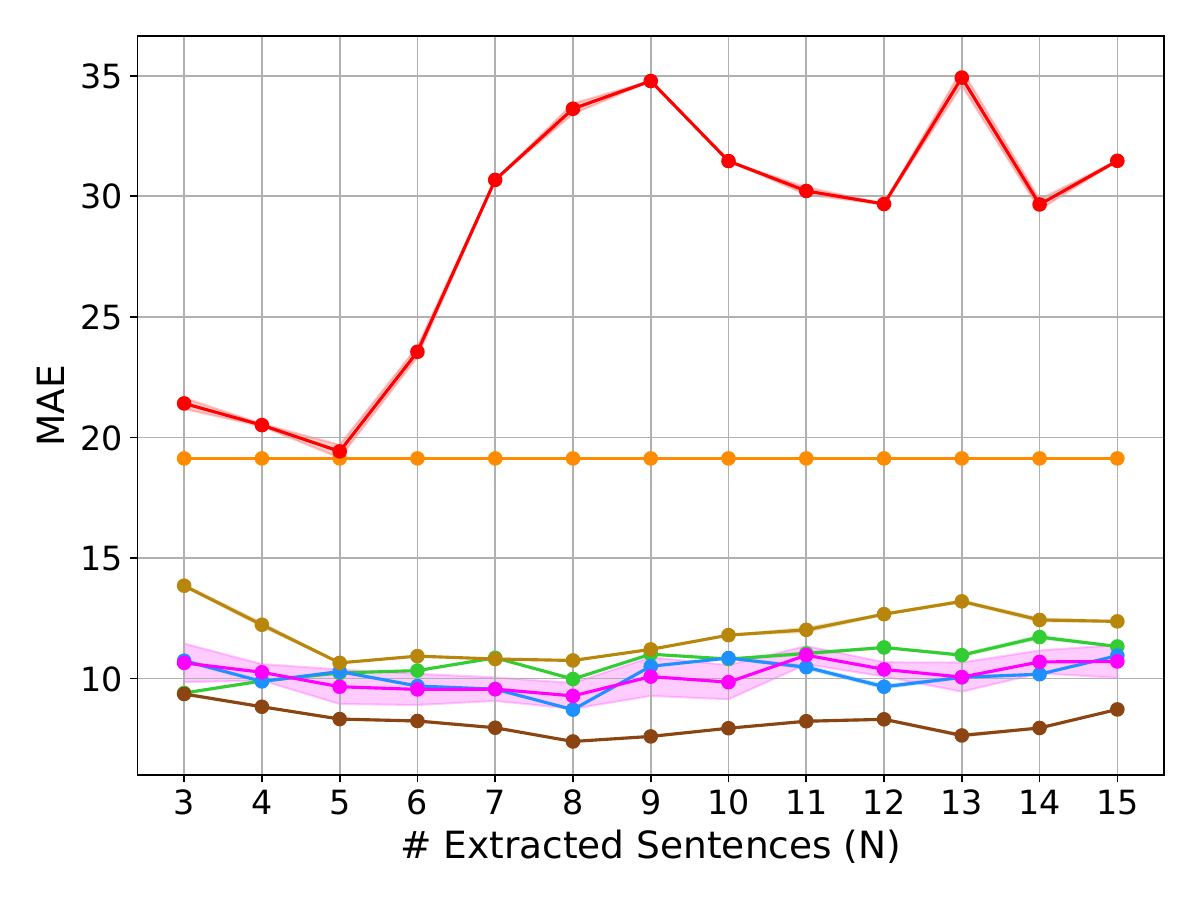}
        \vspace{-1.75em}
        \caption{Mistral 7b Instruct MAE Curve}
        \label{fig: Avg MAE mistral7b-gamespot}
        \vspace{1em}
    \end{subfigure}
    \hfill
    \begin{subfigure}[b]{0.328\textwidth}
        \centering
        \includegraphics[width=1.05\textwidth]{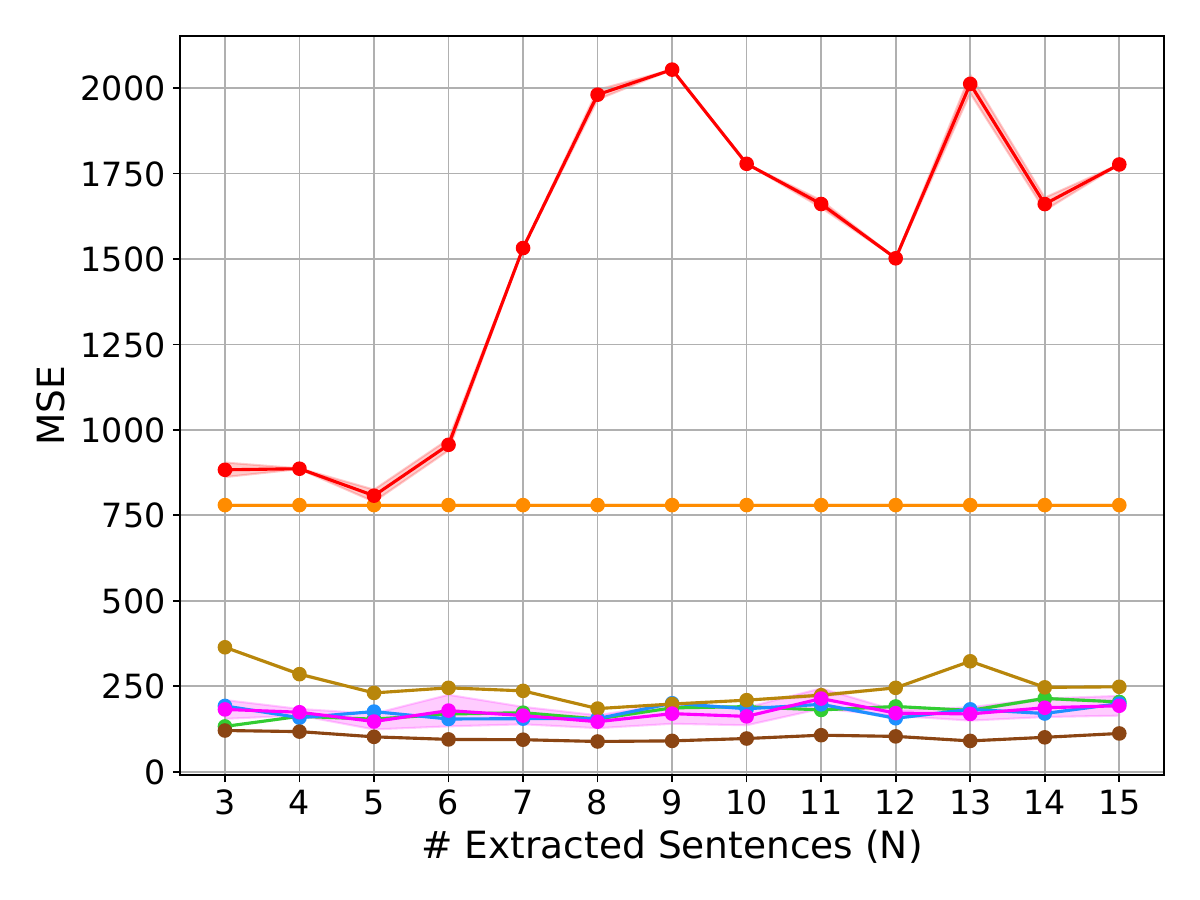}
        \vspace{-1.75em}
        \caption{Mistral 7b Instruct MSE Curve}
        \label{fig: Avg MSE mistral7b-gamespot}
        \vspace{1em}
    \end{subfigure}

    \vspace{-0.4em}

    \begin{subfigure}[b]{0.328\textwidth}
        \centering
        \includegraphics[width=1.05\textwidth]{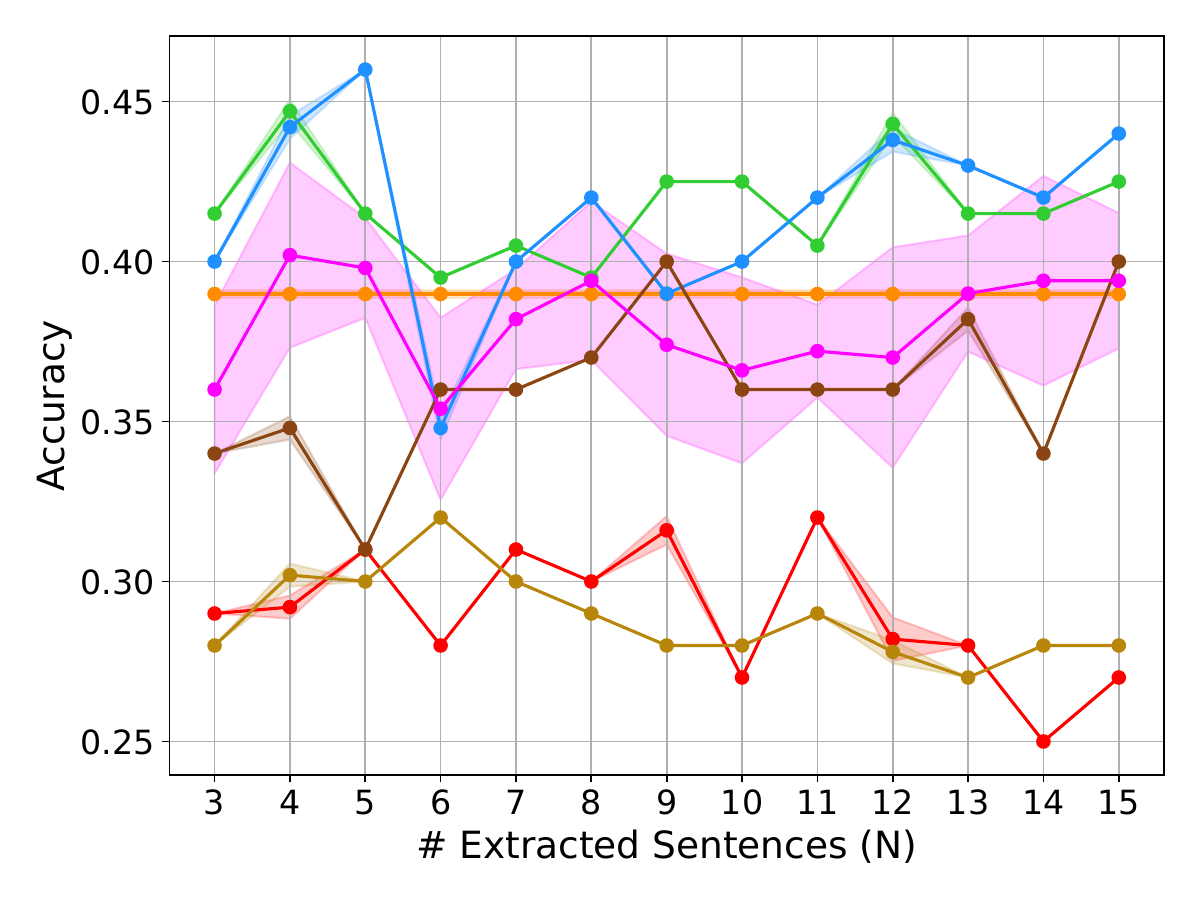}
        \vspace{-1.75em}
        \caption{Llama3 8b Instruct Accuracy Curve}
        \label{fig: Avg Acc llama3-gamespot}
        \vspace{0.3em}
    \end{subfigure}
    \hfill
    \begin{subfigure}[b]{0.328\textwidth}
        \centering
        \includegraphics[width=1.05\textwidth]{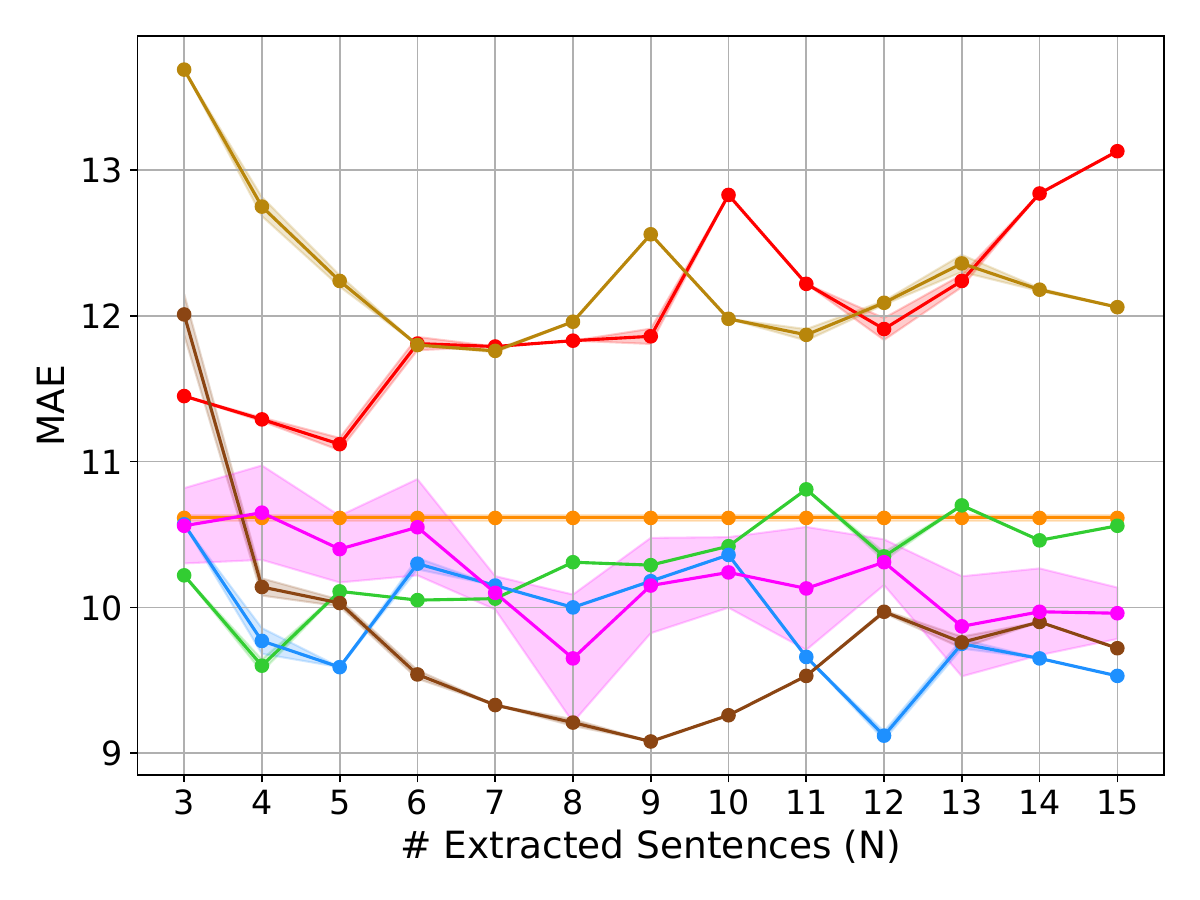}
        \vspace{-1.75em}
        \caption{Llama3 8b Instruct MAE Curve}
        \label{fig: Avg MAE llama3-gamespot}
        \vspace{0.3em}
    \end{subfigure}
    \hfill
    \begin{subfigure}[b]{0.328\textwidth}
        \centering
        \includegraphics[width=1.05\textwidth]{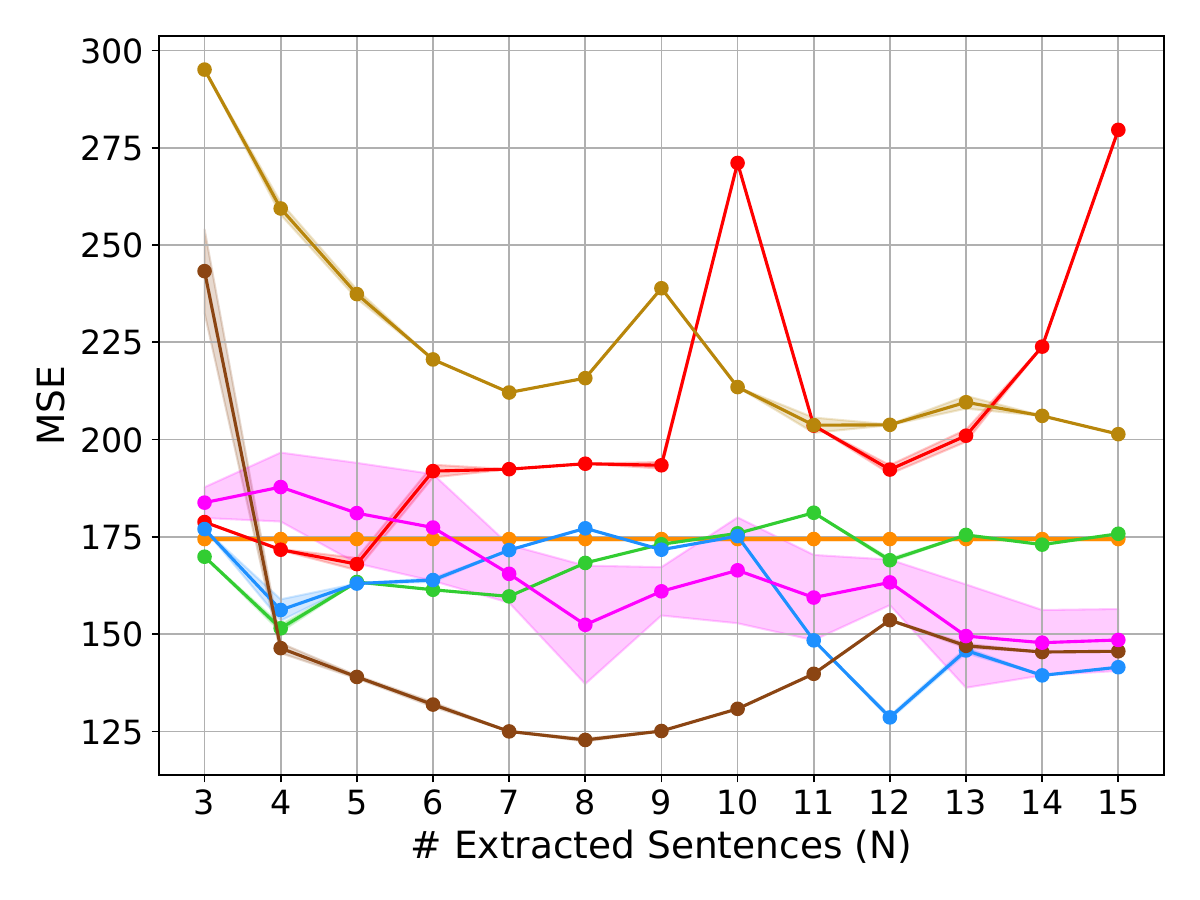}
        \vspace{-1.75em}
        \caption{Llama3 8b Instruct MSE Curve}
        \label{fig: Avg MSE llama3-gamespot}
        \vspace{0.3em}
    \end{subfigure}
    \vspace{-2em}
    \caption{\small LLMs performance over 5 runs on sentiment analysis on GameSpot dataset. The ``Full Text'' scenario is a horizontal line since it always contains the full text, not summary/truncated text and we have included it as a horizontal line here as baseline}
    \label{fig: Avg performance gamespot}
\end{figure*}

\end{document}